\pdfoutput=1
\documentclass[runningheads,a4paper]{article}

\usepackage{paralist}
\usepackage{amsmath}
\usepackage{graphicx}
\usepackage{subfig}
\usepackage[belowskip=-12pt,aboveskip=3pt]{caption}
\usepackage{booktabs}
\usepackage{longtable}
\usepackage{tabularx}
\usepackage{multirow}
\usepackage{textcomp}
\usepackage[table,x11names]{xcolor}
\usepackage{authblk}
\usepackage{url}

\usepackage{filecontents}

\graphicspath{{./img/}}

\begin{document}

\title{Automatic View-Point Selection for Inter-Operative Endoscopic Surveillance}
\author[1,3]{A.S. Vemuri}
\author[3]{S.A. Nicolau}
\author[3]{J. Marescaux}
\author[3]{L. Soler}
\author[2]{N. Ayache}
\affil[1]{IHU, 1 Place de l'Hopital, 67091 Strasbourg Cedex, FRANCE}
\affil[2]{INRIA Sophia Antipolis, 06902 Sophia Antipolis Cedex, FRANCE}
\affil[3]{IRCAD, 1 Place de l'Hopital, 67091 Strasbourg Cedex, FRANCE}

\date{}

\renewcommand\Authands{ and }

\maketitle

\begin{abstract}

Esophageal adenocarcinoma arises from Barrett's esophagus, which is the most serious complication of gastroesophageal reflux disease. Strategies for screening involve periodic surveillance and tissue biopsies. A major challenge in such regular examinations is to record and track the disease evolution and re-localization of biopsied sites to provide targeted treatments. In this paper, we extend our original inter-operative relocalization framework to provide a constrained image based search for obtaining the best view-point match to the live view. Within this context we investigate the effect of, \begin{inparaenum}[\upshape(a\upshape)]
\item the choice of feature descriptors and color-space, 
\item filtering of uninformative frames,
\item endoscopic modality,
\end{inparaenum} for view-point localization. Our experiments indicate an improvement in the best view-point retrieval rate to $[92\%,87\%]$ from $[73\%,76\%]$ (in our previous approach) for NBI and WL.
\end{abstract}

\section{Introduction}
\label{sec:introduction}

The incidence of esophageal adenocarcinoma has risen dramatically over the past three decades in western countries. Adenocarcinoma of the esophagus appears to arise from the Barrett's muscosa through progressive degrees of dysplasia \cite{Conteduca2012}. The guidelines \cite{Wang2008c} prescribe different levels of surveillance intervals depending on the degree of dysplasia and the cost-effectiveness. One of the key challenges to regular surveillance is re-localization of biopsy sites between two surveillance procedures. 

\begin{figure}[!htb]
	\centering
	\includegraphics[width=0.7\linewidth]{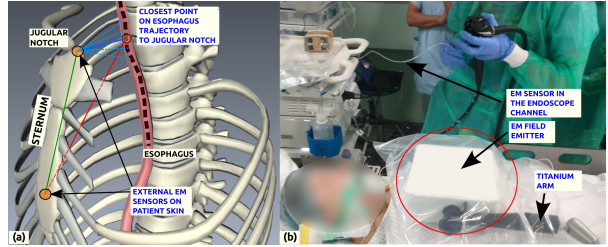}
	\caption{System setup}
	\label{fig:system}
\end{figure}

In an earlier work \cite{Vemuri2013} we provided a first approach to inter-operative video sychronization using an electromagnetic tracker (EMTS). Fig. (\ref{fig:system}) shows the system setup. In our framework we used an EM sensor inside the endoscope channel to track its position inside the esophagus and two external sensors for providing the anatomical landmarks on the patient. We performed simultaneous capture from EMTS and the endoscopic frame to generate a database where, each captured image has a corresponding 3D position associated with it. Firstly we performed inter-operative registration using external sensors on the patient. Then to provide inter-operative video synchronization, we used the 3D position obtained from the EM sensor to determine the nearest neighbour (EMNN). The corresponding image for the EMNN provided the localized view in the esophagus. We define this as gross-localization in the context of our problem. From the clinical point of view, this can be applicable to the gastroenterologist (GIS) in two ways. \begin{inparaenum}[(i)]
\item \emph{Differential Surveillance} (DS): By juxtaposing the synchronized view with the live view, it provides the GIS, the corresponding variations at the same location in the esophagus.
\item \emph{Biopsy Site Relocalization} (BSR): In an alternate scenario, when only the biopsy sites are stored in the database, the 3D position of the sensor in the endoscope would allow the GIS for relocalizing the biopsy sites.
\end{inparaenum} 



\begin{figure}[!htb]
      \centering
      \captionsetup[subfigure]{justification=centering}
      \subfloat[]{\label{sf:query}\includegraphics[width=0.14\linewidth]{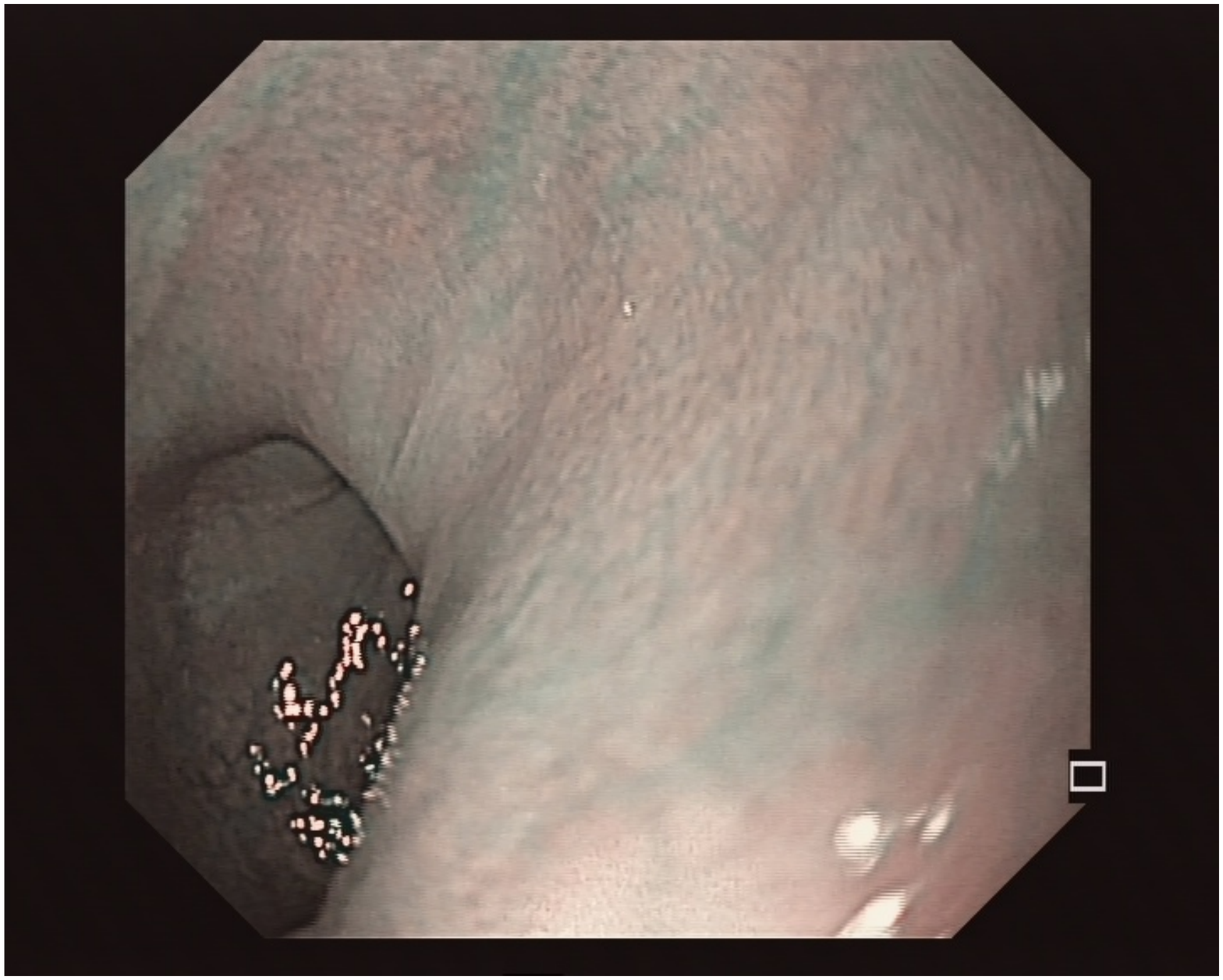}}\quad      
      \subfloat[]{\label{sf:M1}\includegraphics[width=0.14\linewidth]{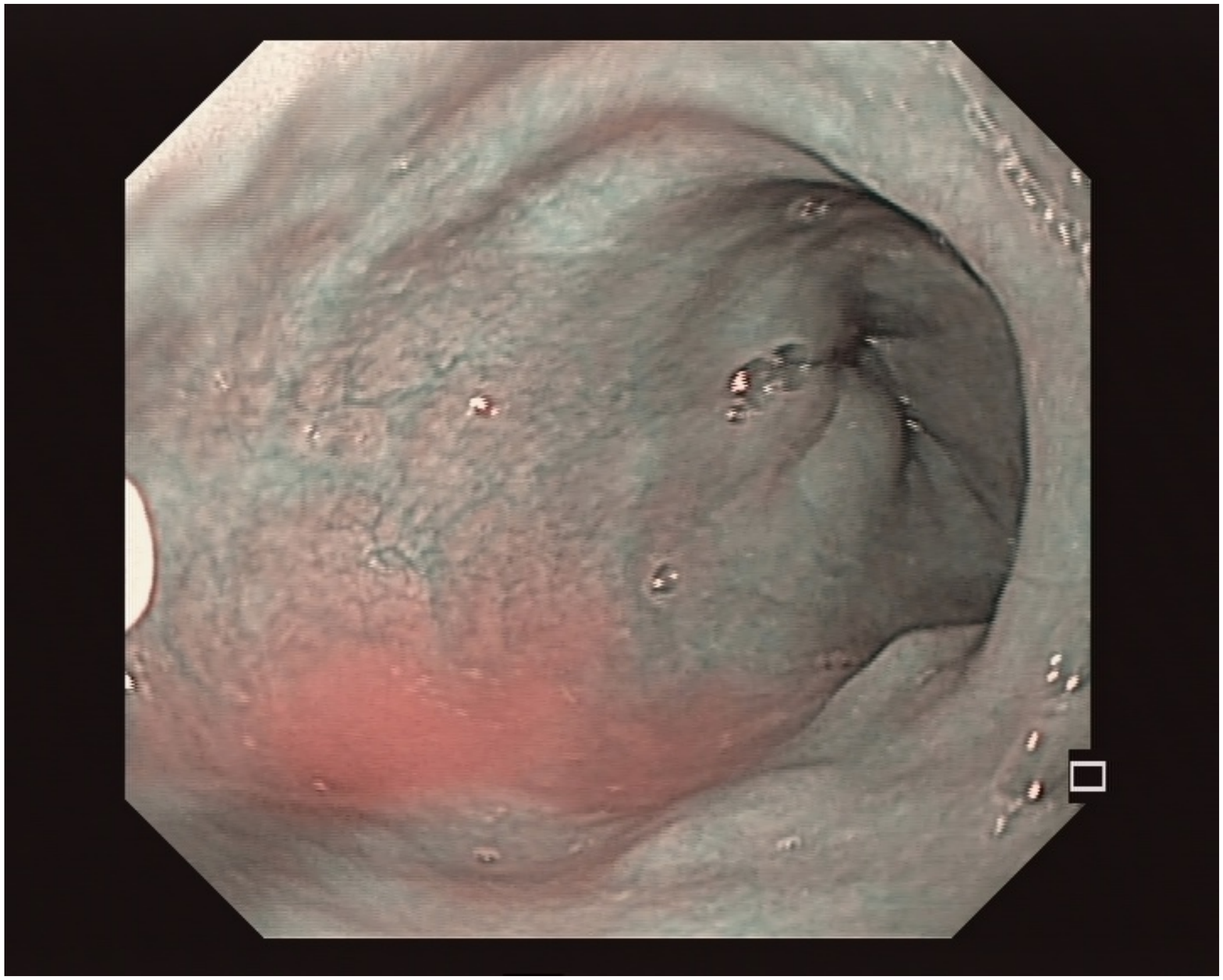}}\quad
      \subfloat[]{\label{sf:M2}\includegraphics[width=0.14\linewidth]{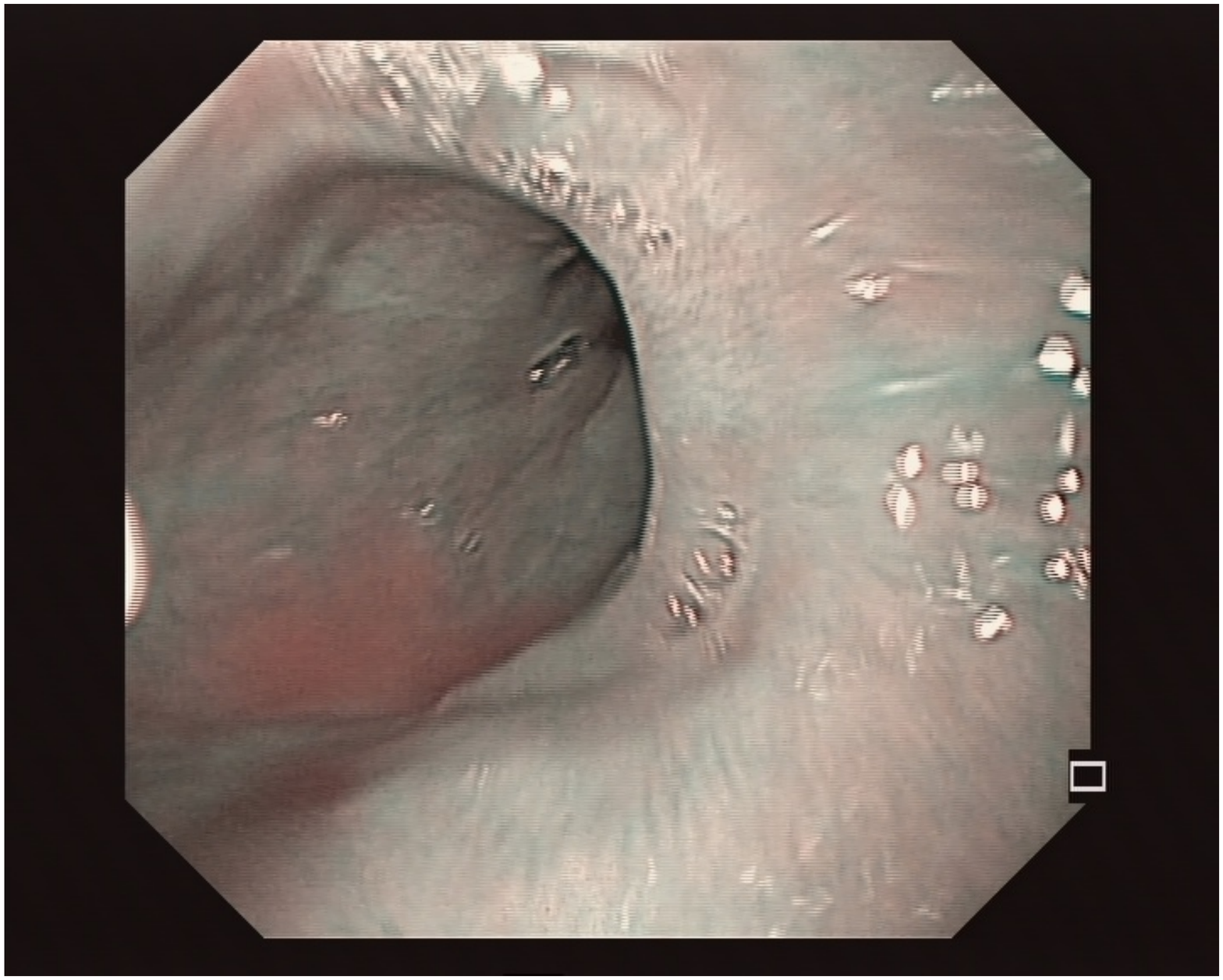}}\quad
      \subfloat[]{\label{sf:M3}\includegraphics[width=0.14\linewidth]{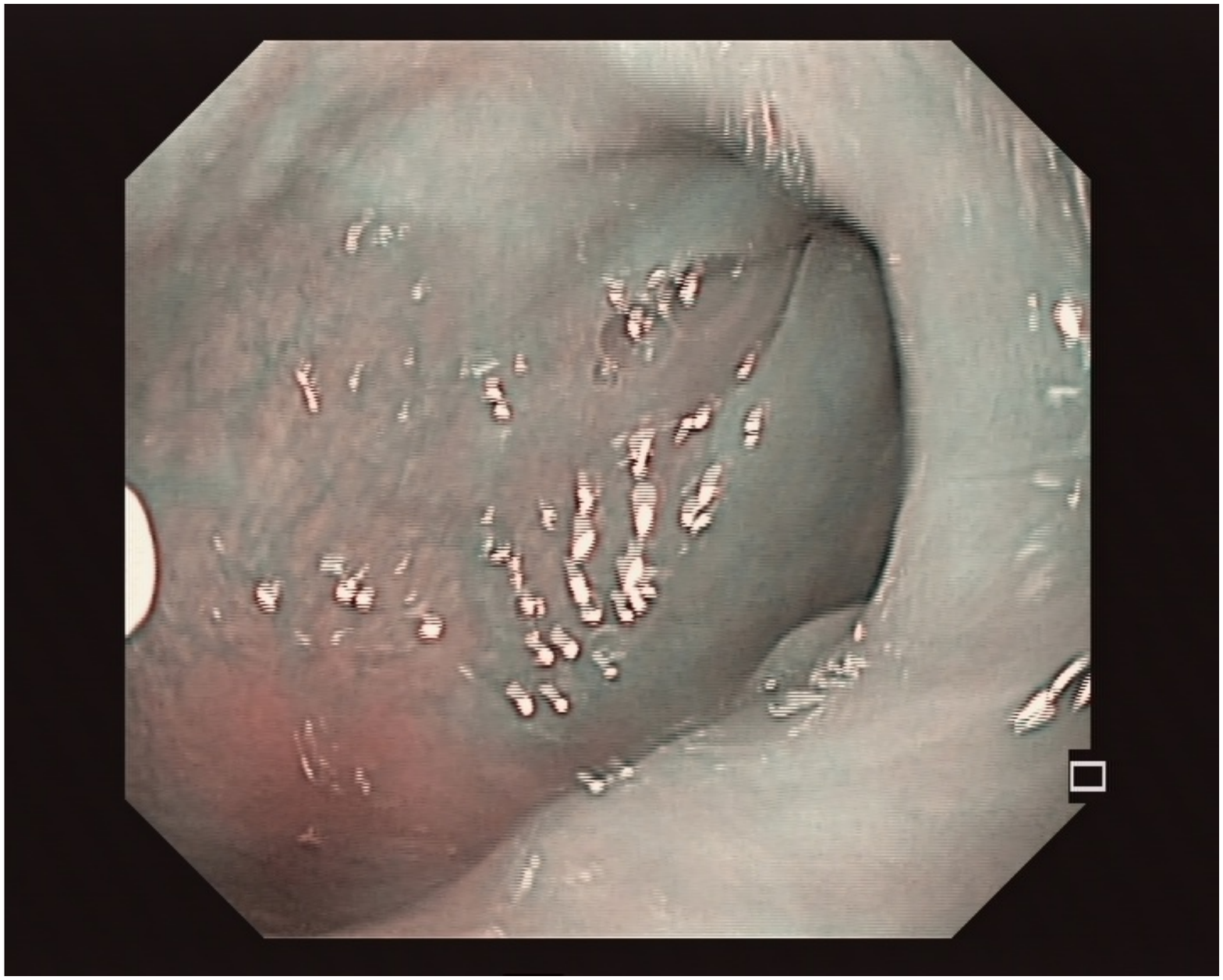}}\quad
      \subfloat[]{\label{sf:M4}\includegraphics[width=0.14\linewidth]{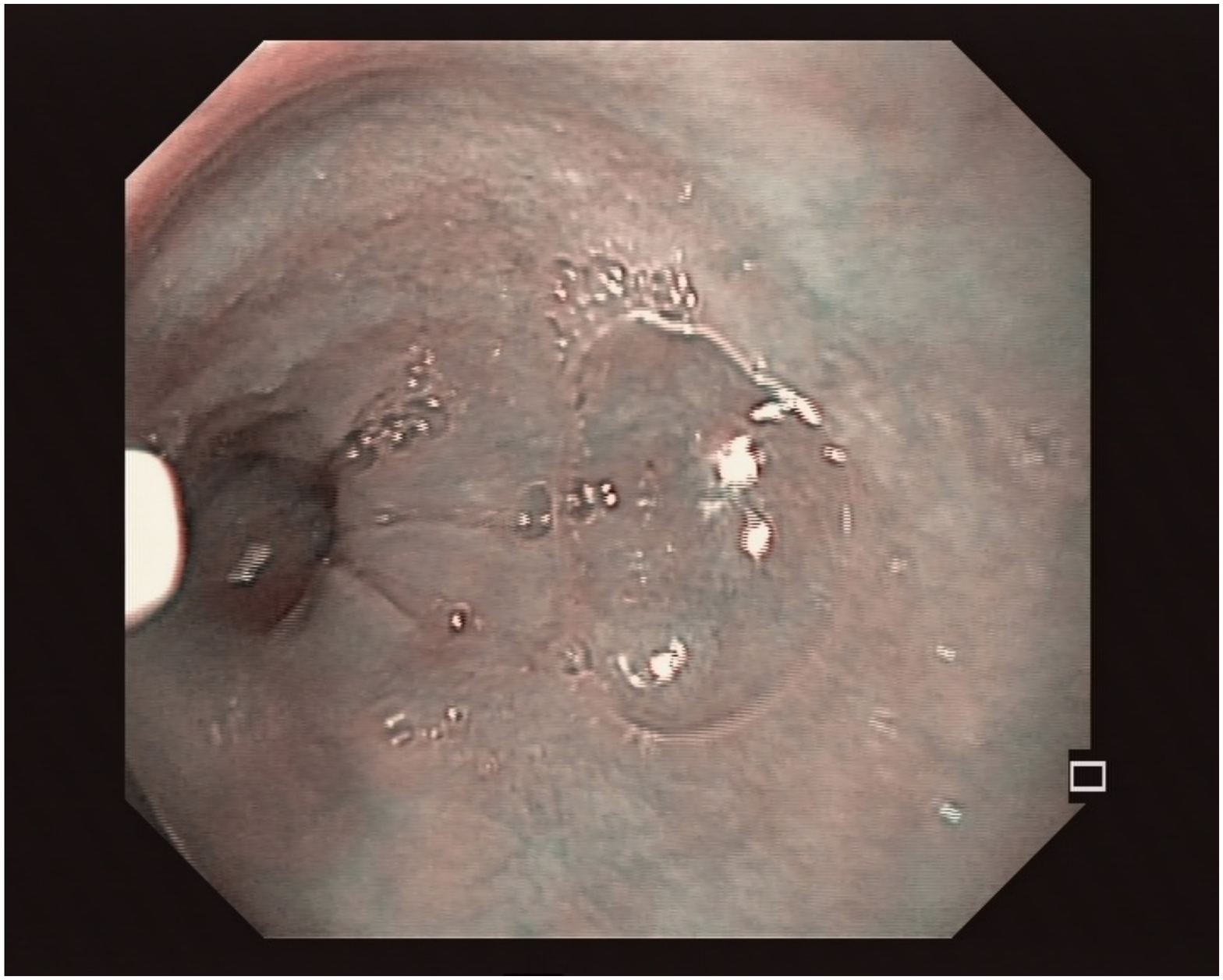}}\quad
      \subfloat[]{\label{sf:M5}\includegraphics[width=0.14\linewidth]{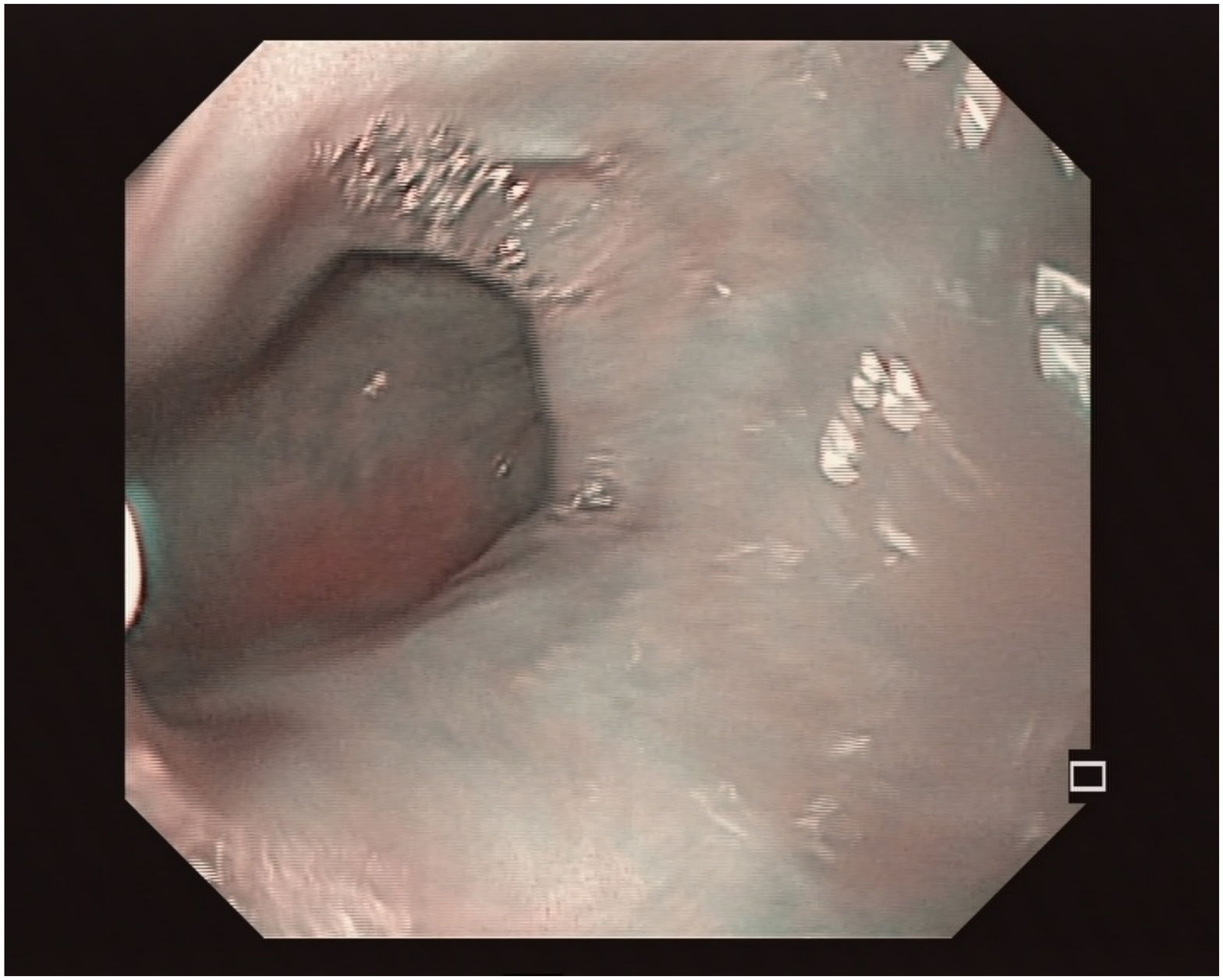}}\quad
      \subfloat[\scriptsize{Blurred}]{\label{sf:UF1}\includegraphics[width=0.175\linewidth]{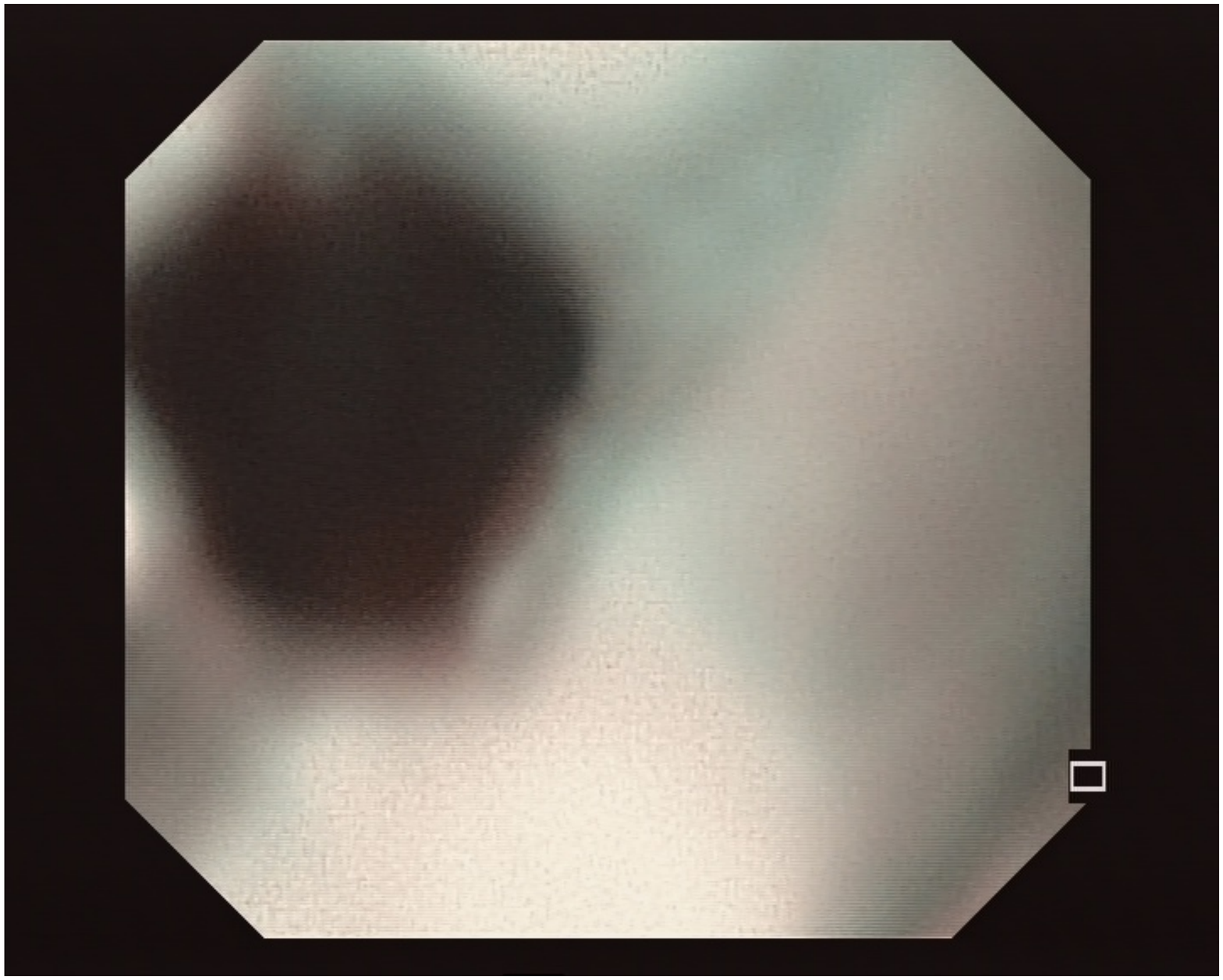}}\quad      
      \subfloat[\scriptsize{Contact}]{\label{sf:UF2}\includegraphics[width=0.175\linewidth]{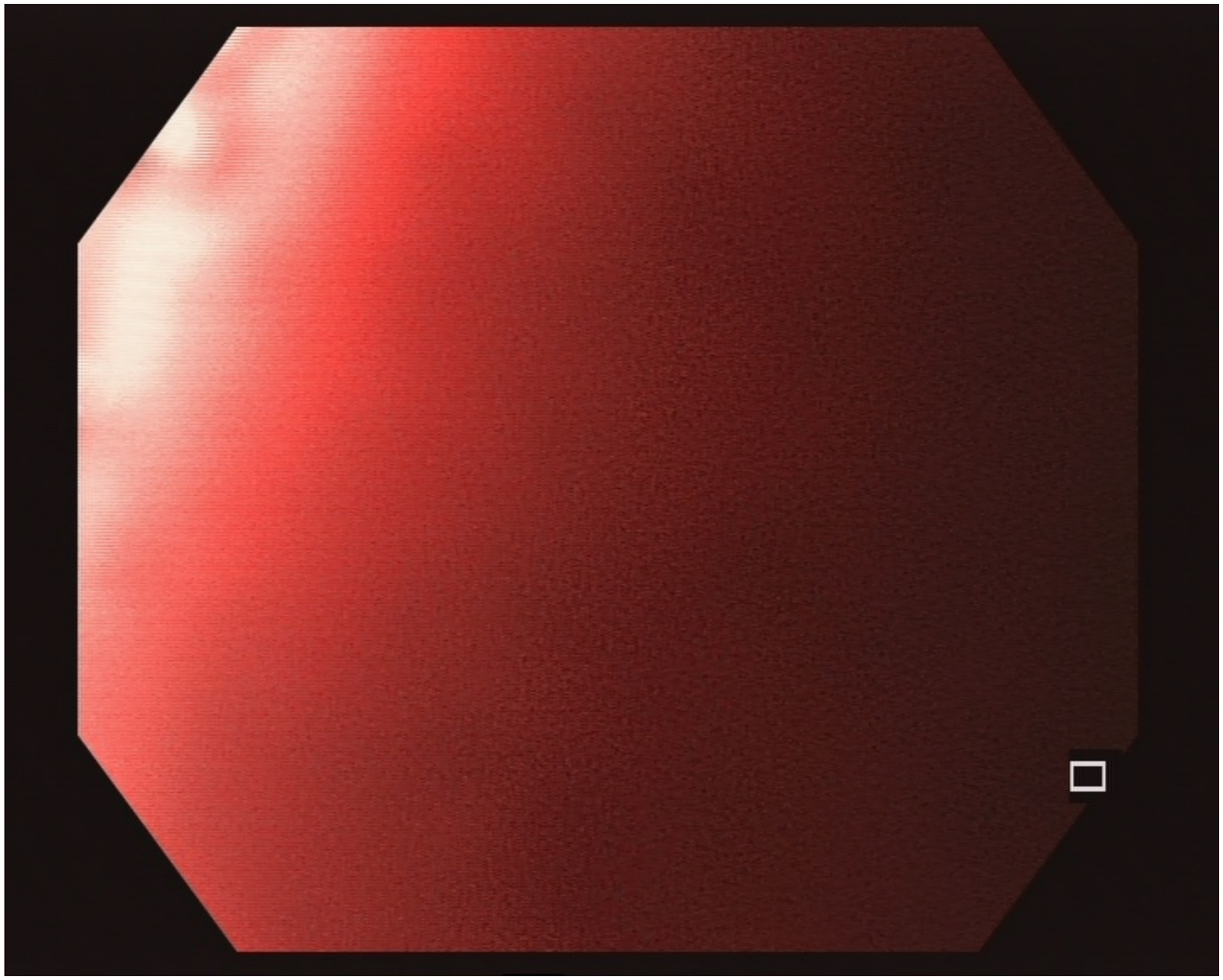}}\quad
      \subfloat[\scriptsize{Motion-Blur}]{\label{sf:UF3}\includegraphics[width=0.175\linewidth]{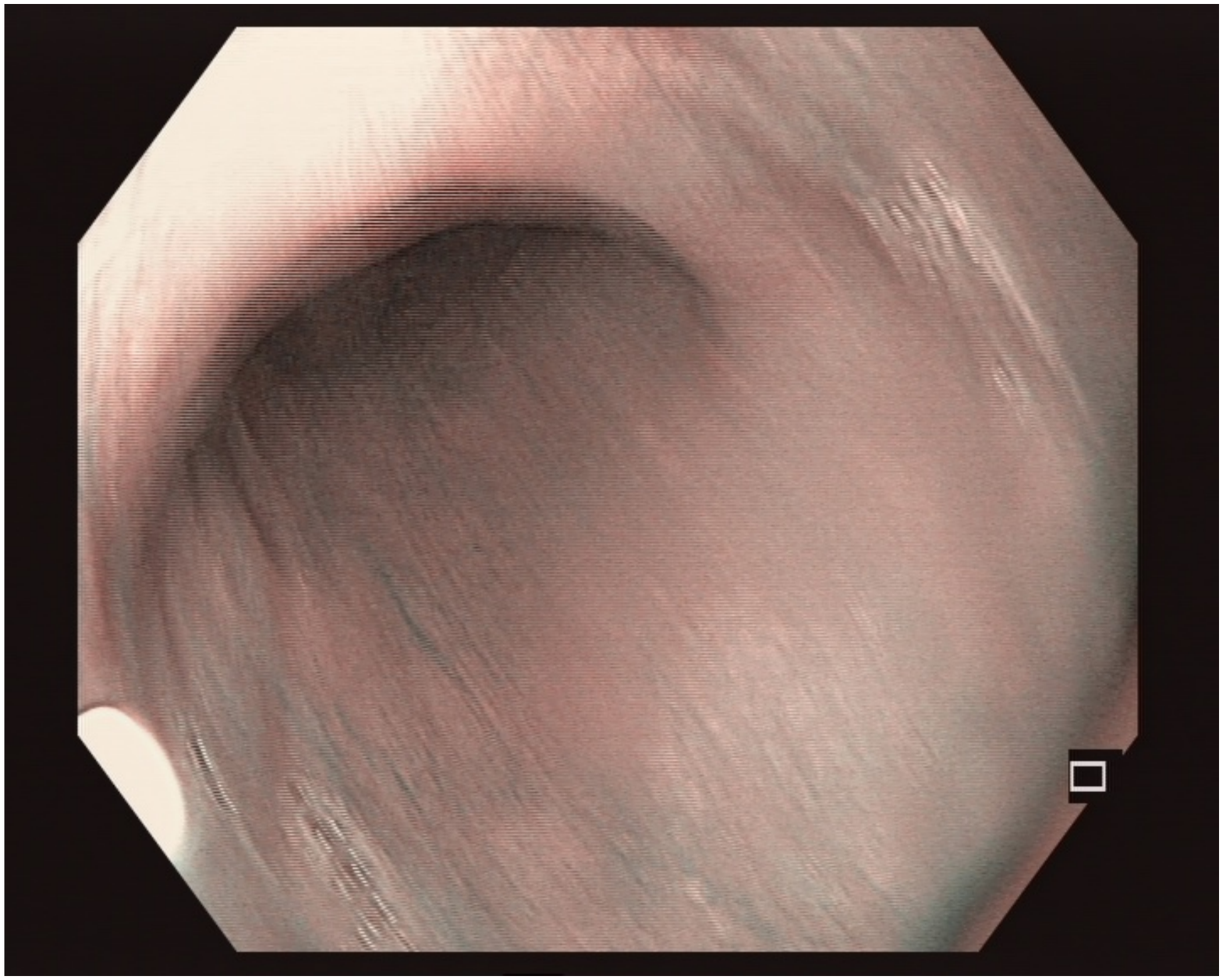}}\quad
      \subfloat[\scriptsize{Contraction}]{\label{sf:UF4}\includegraphics[width=0.175\linewidth]{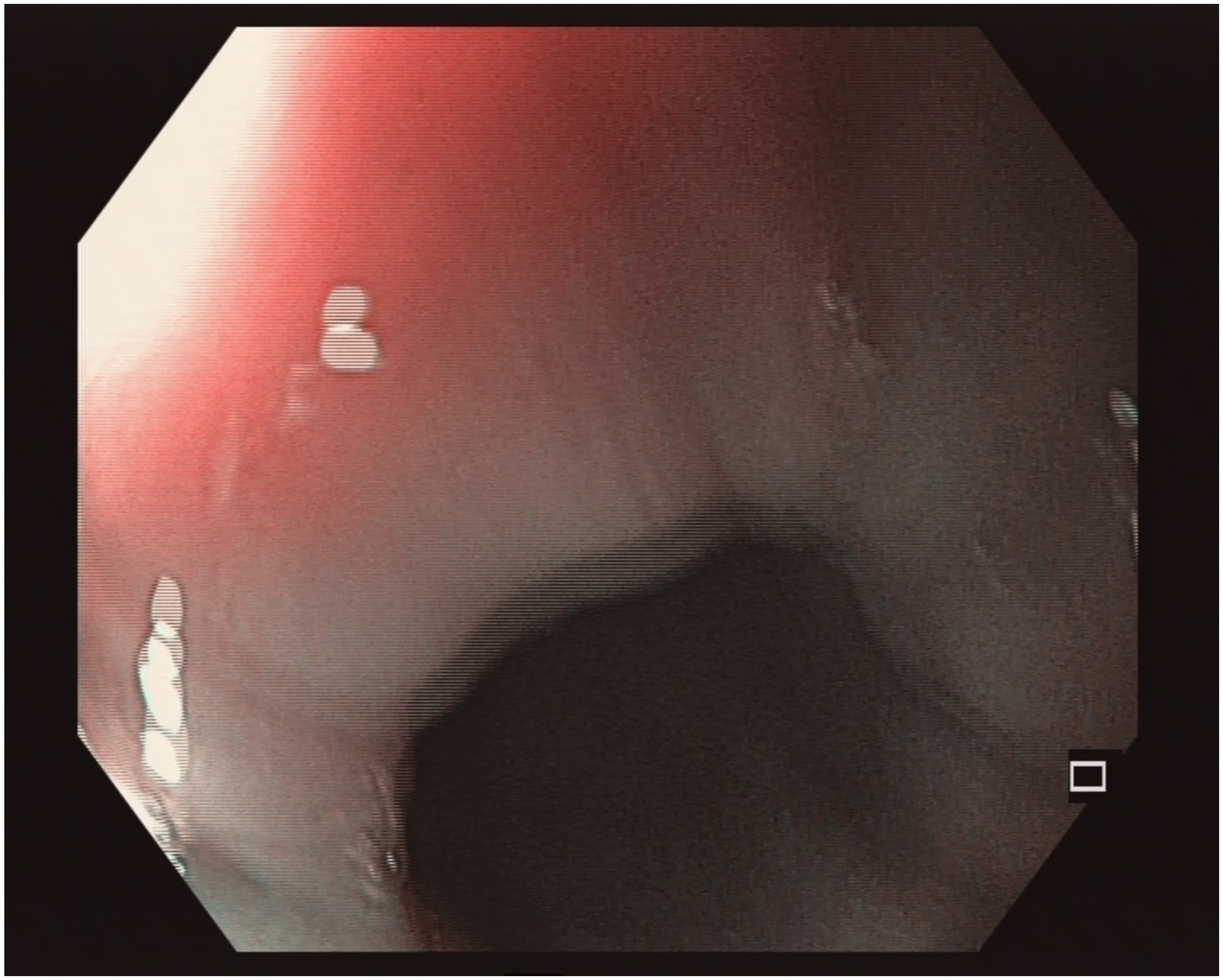}}\quad
      \subfloat[\scriptsize{Fluid}]{\label{sf:UF5}\includegraphics[width=0.175\linewidth]{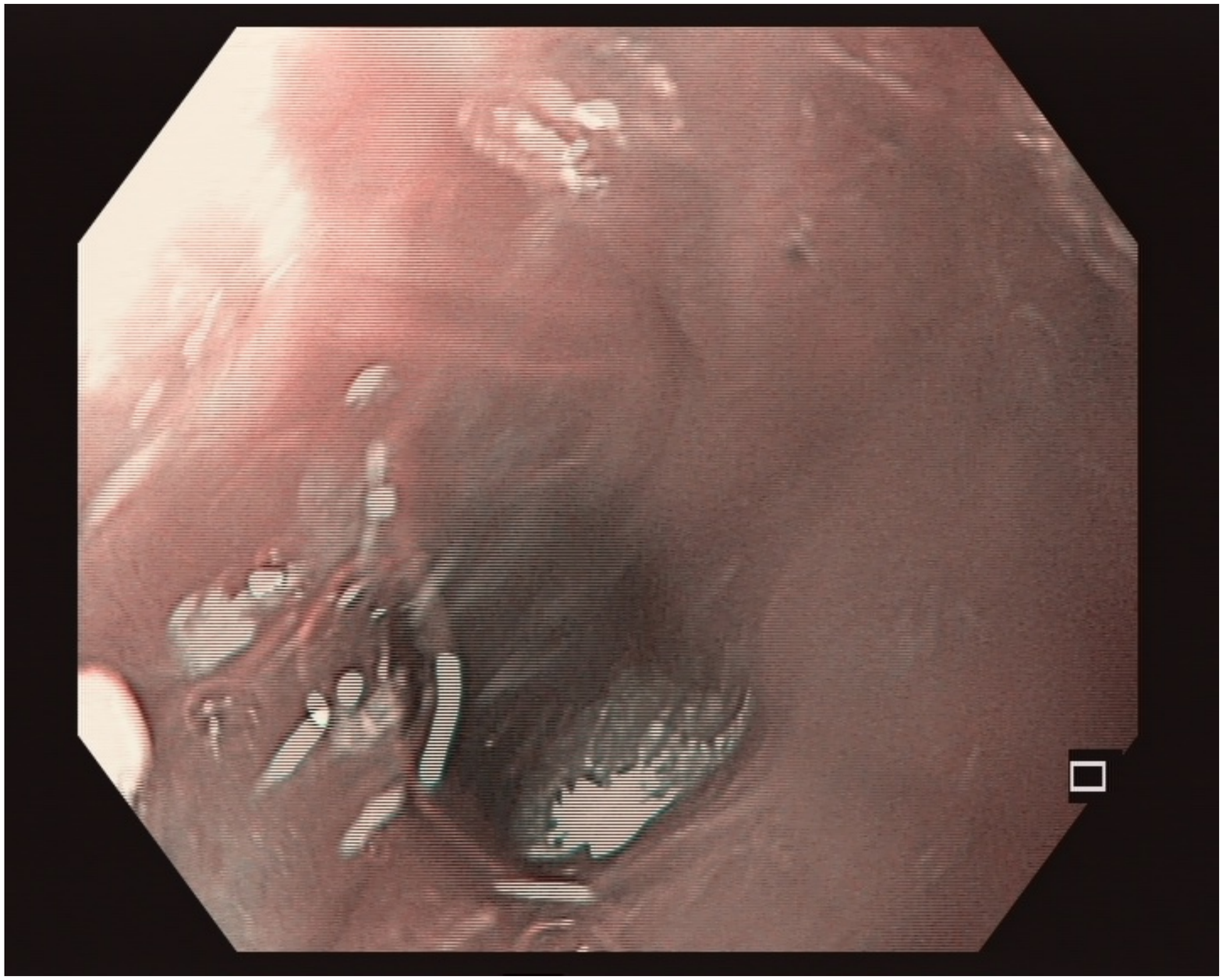}}\quad
      \caption{\emph{k}-EMNN matches. \protect\subref{sf:query}: Query, \protect\subref{sf:M1}-\protect\subref{sf:M5}: matched frames according to EMTS with scores: $[0,1,0,2,1]$ as assinged by the expert section. \ref{sec:experiments}. \protect\subref{sf:UF1}-\protect\subref{sf:UF5} present the sample UI frames.\\}
      \label{fig:wrong_matches_UI_frames}
\end{figure}

The problem of BSR has been addressed in our earlier work \cite{Vemuri2015}. The goal of this paper is to provide a constrained approach to scene association for DS. The EMNN does not necessarily provide the best view-point from GIS's perspective. This could be due to a combination of three reasons: \begin{inparaenum}[\itshape a\upshape)]
  \item The matched view could vary from the live view in the esophagus fig. \ref{fig:wrong_matches_UI_frames}\protect\subref{sf:M1}-\ref{fig:wrong_matches_UI_frames}\protect\subref{sf:M5},
  \item The matched view could be an uninformative (UI) frame as shown in fig. \ref{fig:wrong_matches_UI_frames}\protect\subref{sf:UF1}-\ref{fig:wrong_matches_UI_frames}\protect\subref{sf:UF5}.
  \item In \cite{Vemuri2015} we have evaluated the influence of uncertainty of the placement the external markers. Essentially, it leads to an average depth estimation error along the esophagus of $\pm 10mm$ for a $95\%$ confidence interval. 
\end{inparaenum}
It is thus important to provide an intelligent selection of the best matching view using the images in the neighbourhood of the gross-localized region.


\begin{figure}[!htb]
	\centering
	\includegraphics[width=0.91\linewidth]{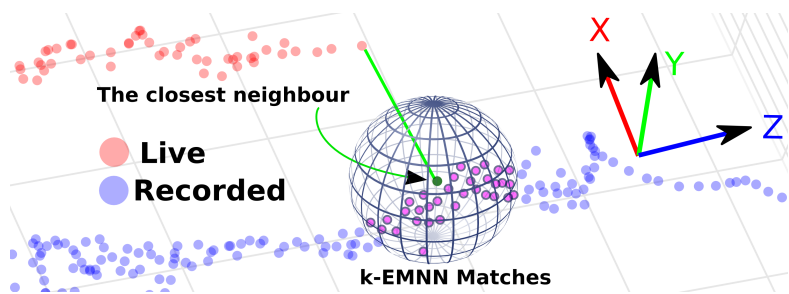}
	\caption{k-Nearest neighbour electromagnetic tracker matches}
	\label{fig:kNNEM}
\end{figure}

We outline the framework of our approach, by classifying the differential surveillance task into three stages:
\begin{enumerate}
 \item \emph{Gross-localization}: Computing the nearest neighbour from the 3D position obtained from EMTS, we obtain an approximate location of the endoscope in the esophagus.
 \item \emph{View-point localization}: Considering the \emph{k}-EMNN matches within a chosen search radius as shown in fig. (\ref{fig:kNNEM}), we extract descriptors to represent the scene and obtain the best matching viewpoint to the live view. The value of \emph{k} varies depending on the number of EMNNs found within the chosen search radius from the closest EMNN. It is important to note that for view-point localization a corresponding matching view that closely resembles the live view may not always be available. However it is important to be able to provide the ``best available'' matching view.
 \item \emph{Inter-frame mapping}: Finally after the best view-point image has been obtained, regions of interest (and tagged biopsy sites) can be mapped from the matched image to the live frame for re-targeting.
\end{enumerate}


This paper focuses on the view-point localization problem which is closely related to scene classification and recognition, that has received considerable attention in computer vision \cite{Van2010}. Endoscopic scene classification has been applied for automatic detection of pathological conditions \cite{Kwitt2012}. In the scope of our work, \cite{Atasoy2012} had proposed training locality preserving projections for low-dimensional embedding of images from a single intervention for localization in subsequent interventions. Our contributions are; firstly, in establishing the need for gross-localization. Secondly, an approach to filtering of UI frames. And thirdly, an evaluation of various descriptor-color-space combinations for narrow band imaging (NBI) and white light (WL) endoscopic modalities (commonly used in clinical practice), to prove the validity of our approach. Section. \ref{subsec:endoscopic_scene_description} presents the various descriptors for endoscopic scene description. Section. \ref{subsec:UIF_removal} presents our approach to removal of UI frames. Section. \ref{sec:experiments} describes the data collection and assignment of ground-truth for both scene recognition and UI frame classification with their results. Section. \ref{sec:discussion_conclusion} lists the important observations.

%
%
%

\section{Methods}
\label{sec:methods}
\subsection{Feature Descriptors and Matching}
\label{subsec:endoscopic_scene_description}
Table \ref{tab:descriptors} presents a summary of various descriptors used in this study. These descriptors were computed in RGB, HSV, Gray-Scale (GS), normalized RGB (norm), log and opponent color-spaces (chosen appropriately for each descriptor). It should be noted that we are not bound by the need for rotation invariance of the descriptors since, in our earlier work \cite{Vemuri2015}, we had established a way to perform orientation correction between matched images using the 6-\emph{dof} EM sensor information. 




\begin{table}[!htp]
    \centering
	\begin{tabular}{lp{9.5cm}}
	    \toprule
	    \centering
	    \emph{ID} & \emph{Descriptor}\\
	    \midrule
	    mLBP\cite{Maenpaa2003} & Multi-scale Local Binary Patterns. At each level the pyramid, the image was divided into non-overlapping cells. A LBP feature vector was computed for each cell which was concatenated into a large feature vector representing the image.\\
	    mHOG\cite{Newell2011} & Multi-scale Histogram of Oriented Gradients. Approach similar to mLBP but with a HOG descriptor for each cell.\\
	    sw-mLBP & Sliding window mLBP. Similar to mLBP, with each cell is a region within a sliding window over the image.\\
	    mLTP\cite{Tan2010} & Multi-scale Local Ternary Patterns.\\
	    mLBP+mHOG & A combined mLBP and mHOG descriptor.\\
	    dSIFT\cite{Lazebnik2006} & Dense scale invariant feature transform. A fast variant computed on non-overlapping cells for each image.\\
	    mLIOP\cite{WangZ2011} & Multi-scale Local Intensity Order Pattern.\\
	    \bottomrule
	\end{tabular}
	\caption{Summary of feature descriptors evaluated in this study. For multi-scale approaches a scale space image pyramid was constructed.}
	\label{tab:descriptors}
\end{table}



\subsection{Uninformative Frame Removal}
\label{subsec:UIF_removal}
\begin{figure}[!htp]
      \centering
      \begin{minipage}[h]{0.45\linewidth}
	      \centering
	      \includegraphics[trim=4.0cm 10cm 3.5cm 10cm,width=\linewidth]{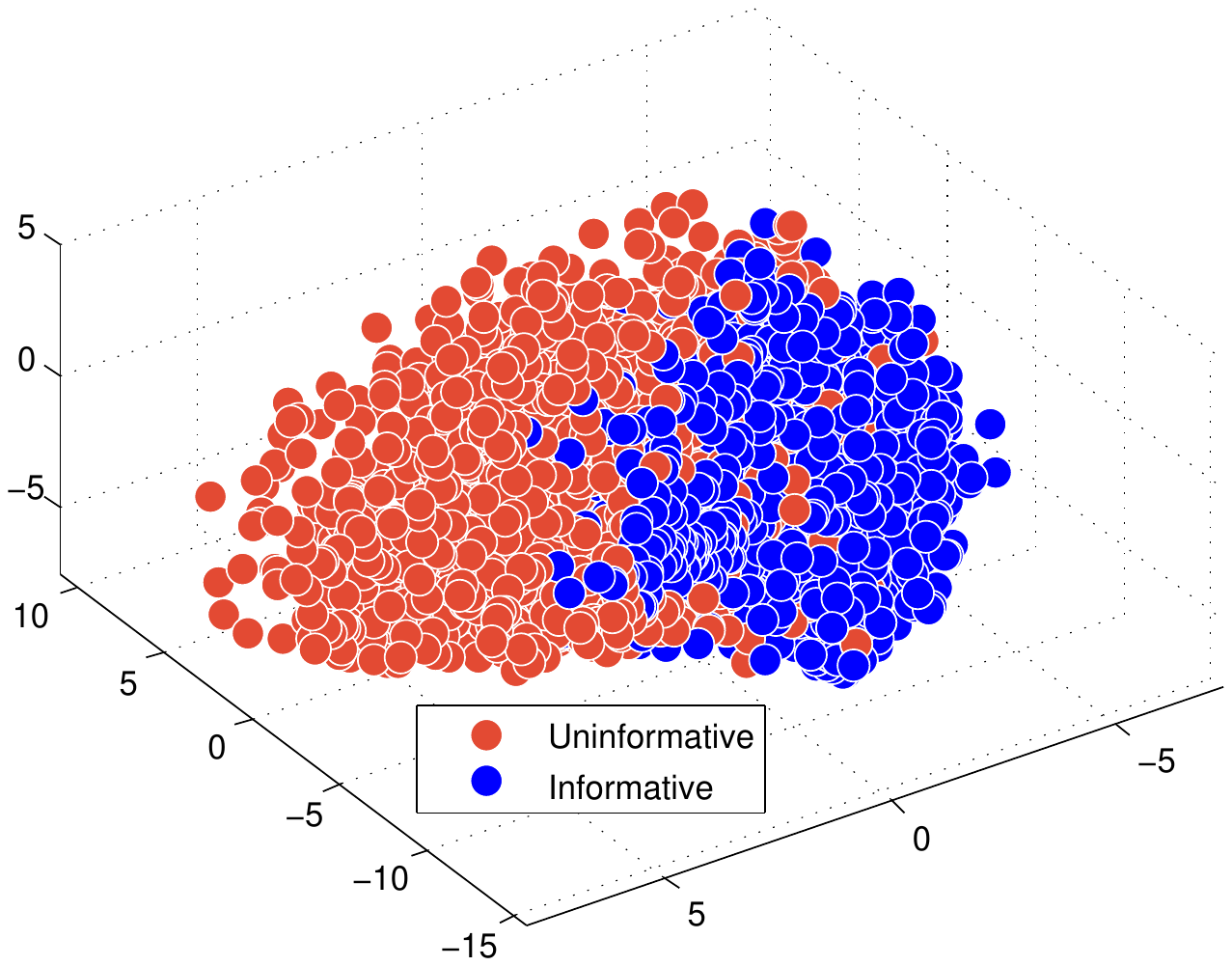}
      \end{minipage}
      \begin{minipage}[h]{0.45\linewidth}
	      \centering
	      \includegraphics[trim=4.0cm 10cm 3.5cm 10cm,width=\linewidth]{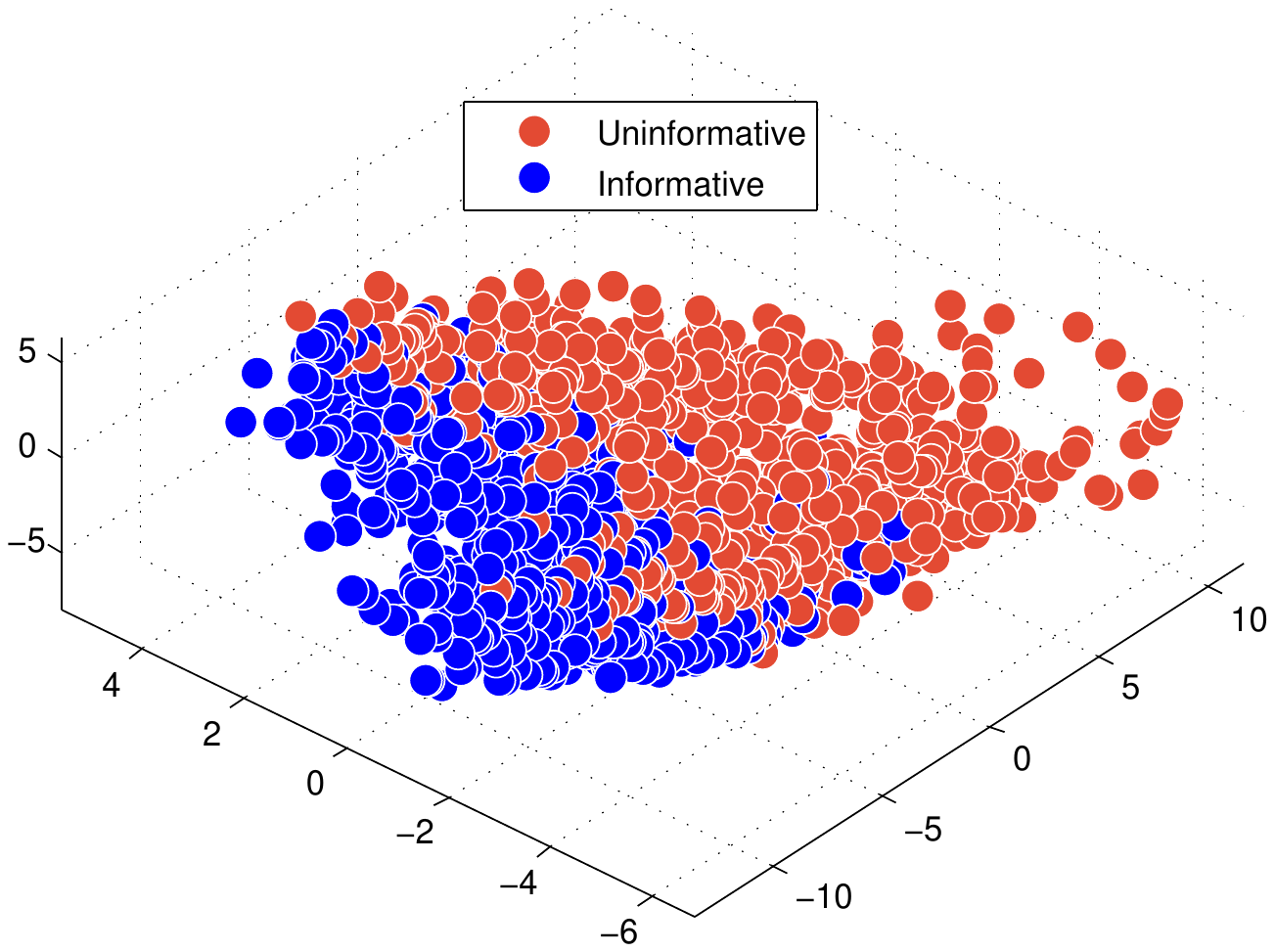}
      \end{minipage}
      \caption{The first three dimensions of the projected feature vectors,  NBI (left) and WL (right), on to the trained basis vectors from PCA.}
      \label{fig:pca_projection}
\end{figure}

A typical endoscopic exploration lasts for several minutes and may include many UI frames fig. \ref{fig:wrong_matches_UI_frames}\protect\subref{sf:UF1}-\ref{fig:wrong_matches_UI_frames}\protect\subref{sf:UF5}. Several approaches have been proposed for removal of UI frames \cite{Bashar2010}. However most have focused on images from capsule endoscopy (CE). Unlike in CE, an endoscopic procedure is shorter, there is insufflation during the procedure and the type of UI frames encountered are different. In our approach, for each image (in GS) we computed a mLBP descriptor. Using the strategy proposed in \cite{Feldman2013} for data selection; \emph{k}-means was applied to obtain the representative cluster centers for the individual classes. Dimensionality reduction was performed for the descriptors from the selected samples using PCA and the resulting feature vectors were used for training. The basis vectors from PCA were then used to project all the image descriptors, shown in fig. (\ref{fig:pca_projection}). An RBF-kernel SVM classifier was trained using LIBSVM \cite{ChangC2011}. First we perfomed leave-one-out cross validation using all the descriptors for parameter selection, where data from \emph{k}-1 interventions was used for training and the $k^{th}$ intervention for testing. These parameters were then used for multiple iterations of data selection (described earlier) to obtain the best model and the average classification rate.

\section{Experiments and Results}
\label{sec:experiments}
We collected data from 7 human subjects, with two surveillance procedures per subject. Between each surveillance procedure the patient underwent gastric treatment, and biopsies were taken for analysis. An Olympus gastroscope was used with WL and NBI modalities. From each recorded trajectory 9 equally spaced query locations were selected to cover about 25cm along the esophagus length. For each selected query locations \emph{k}-EMNN with increasing search radii (10mm-70mm), as shown in fig. (\ref{fig:kNNEM}), were obtained. Hence, for 7 pairs of surveillance endoscopies a total of 63 query locations (each for NBI and WL) were selected. The GIS reviewed the \emph{k}-EMNN matches obtained for each of the search radii and scored the matched images as, 2 - best match, 1 - partial match and 0 - incorrect match. Although these are subjective scores, they help quantify the ideology and the approach to feature based matching in the choice of a good view-point. For each query frame and \emph{k}-EMNN frames, descriptors were computed and matched using chi-squared distance metric. We compared 36 descriptor-color-space combinations. Table \ref{tab:NBI_WL_FilteredScores_SR1} presents results for the 9 best combinations for NBI and WL respectively. The feature descriptors used for this paper are variations of those provided in \cite{Vedaldi08}. An important criteria in selection of these descriptors is also the computation time. Each of these can be applied in a real-time scenario for online classification.

\begin{figure}[!t]
	\begin{center}

		\begin{minipage}[h]{0.17\linewidth}
		 	\centering
		 	\centerline{\includegraphics[width=\linewidth]{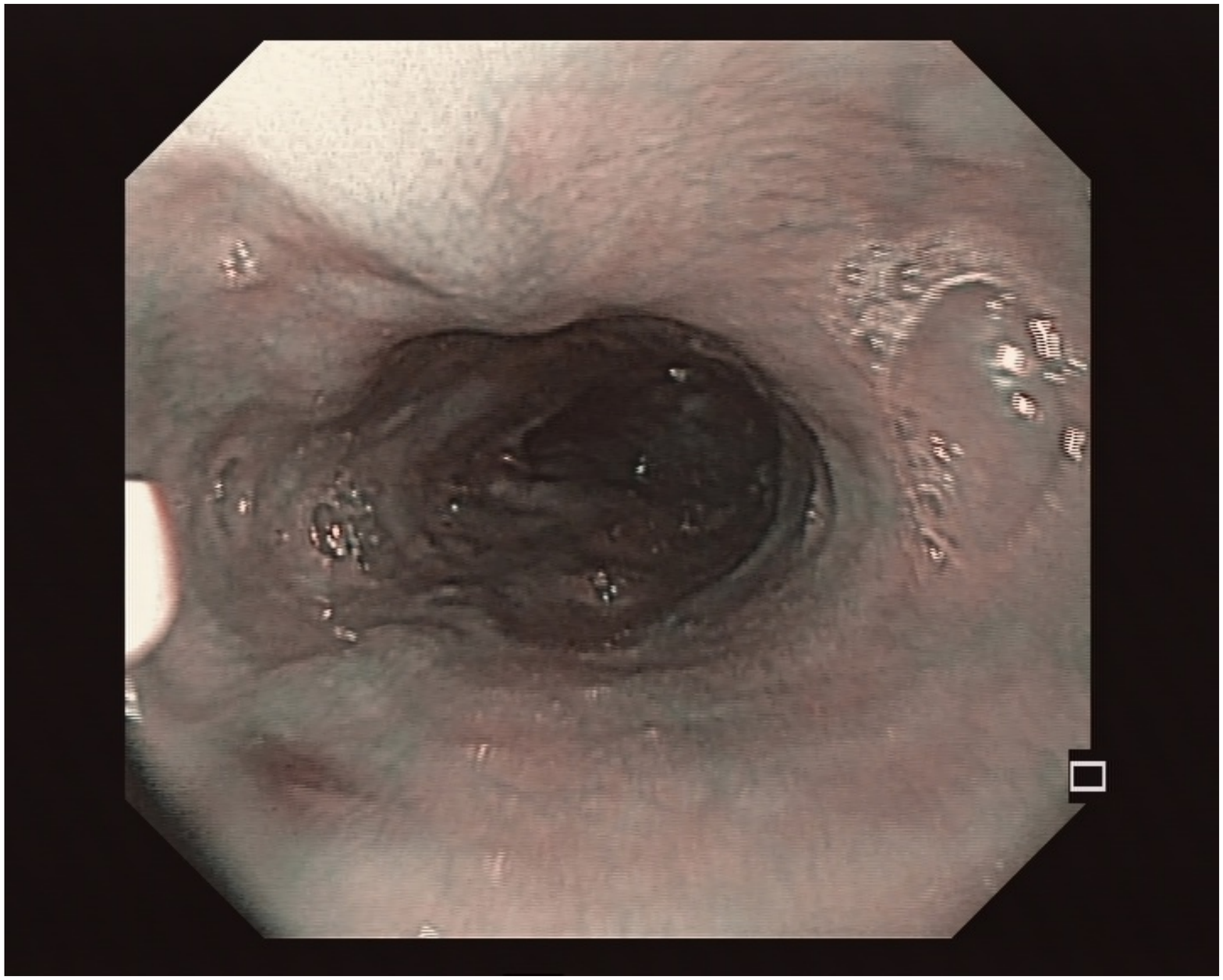}}
		\end{minipage}
		\begin{minipage}[h]{0.17\linewidth}
			 \centering
			 \centerline{\includegraphics[width=\linewidth]{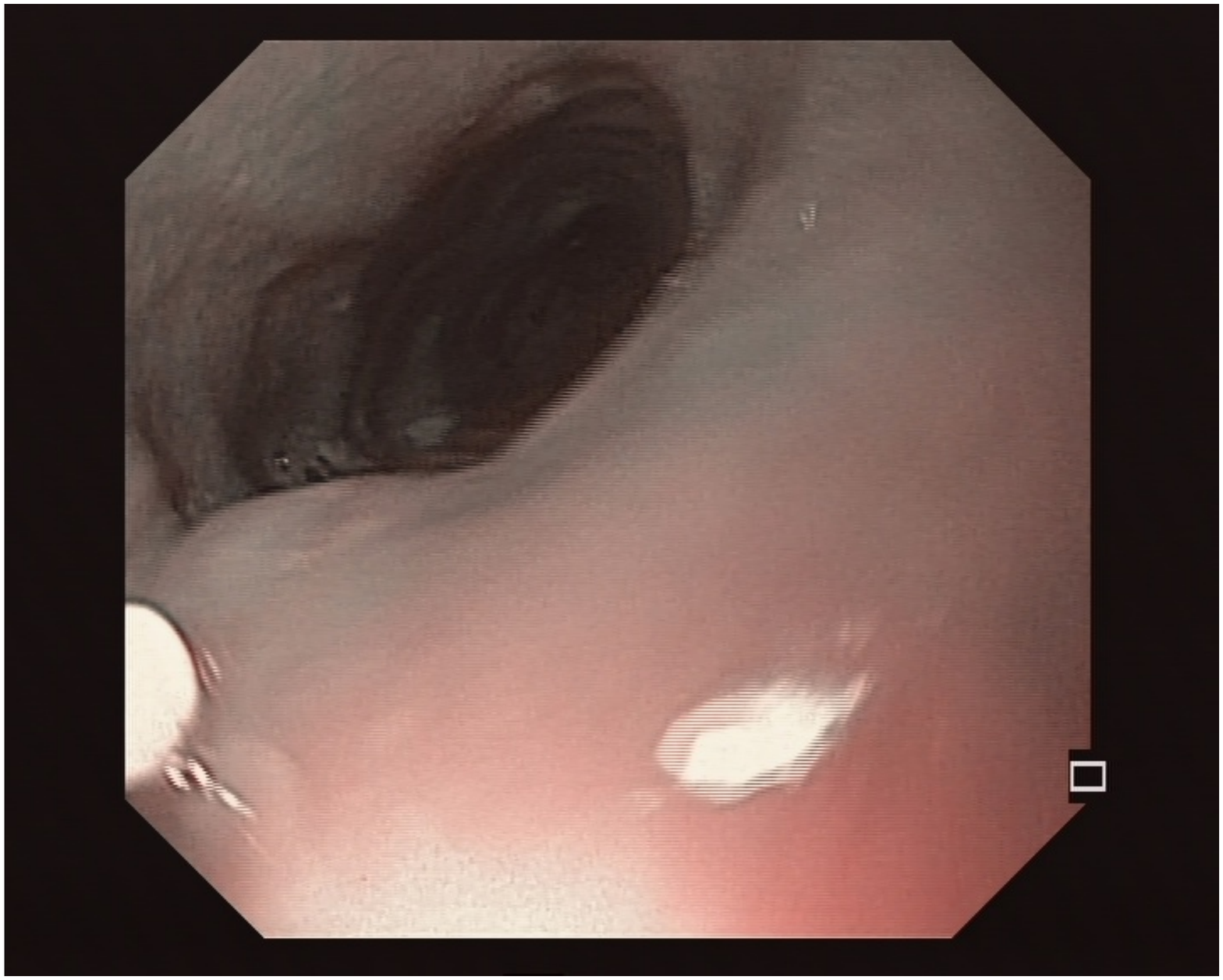}}
		\end{minipage}
		\begin{minipage}[h]{0.17\linewidth}
			 \centering
			 \centerline{\includegraphics[width=\linewidth]{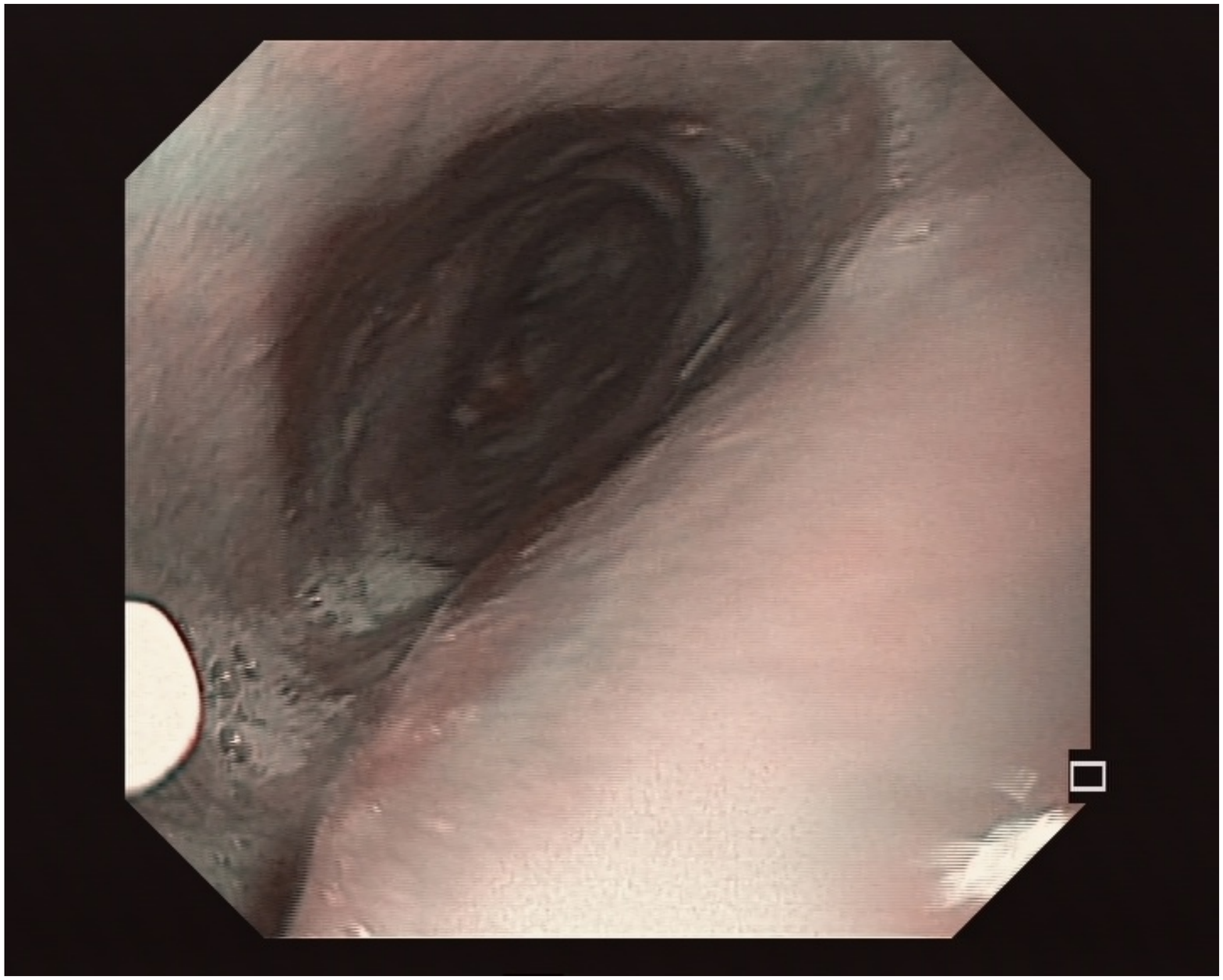}}
		\end{minipage}
		\begin{minipage}[h]{0.17\linewidth}
			 \centering
			 \centerline{\includegraphics[width=\linewidth]{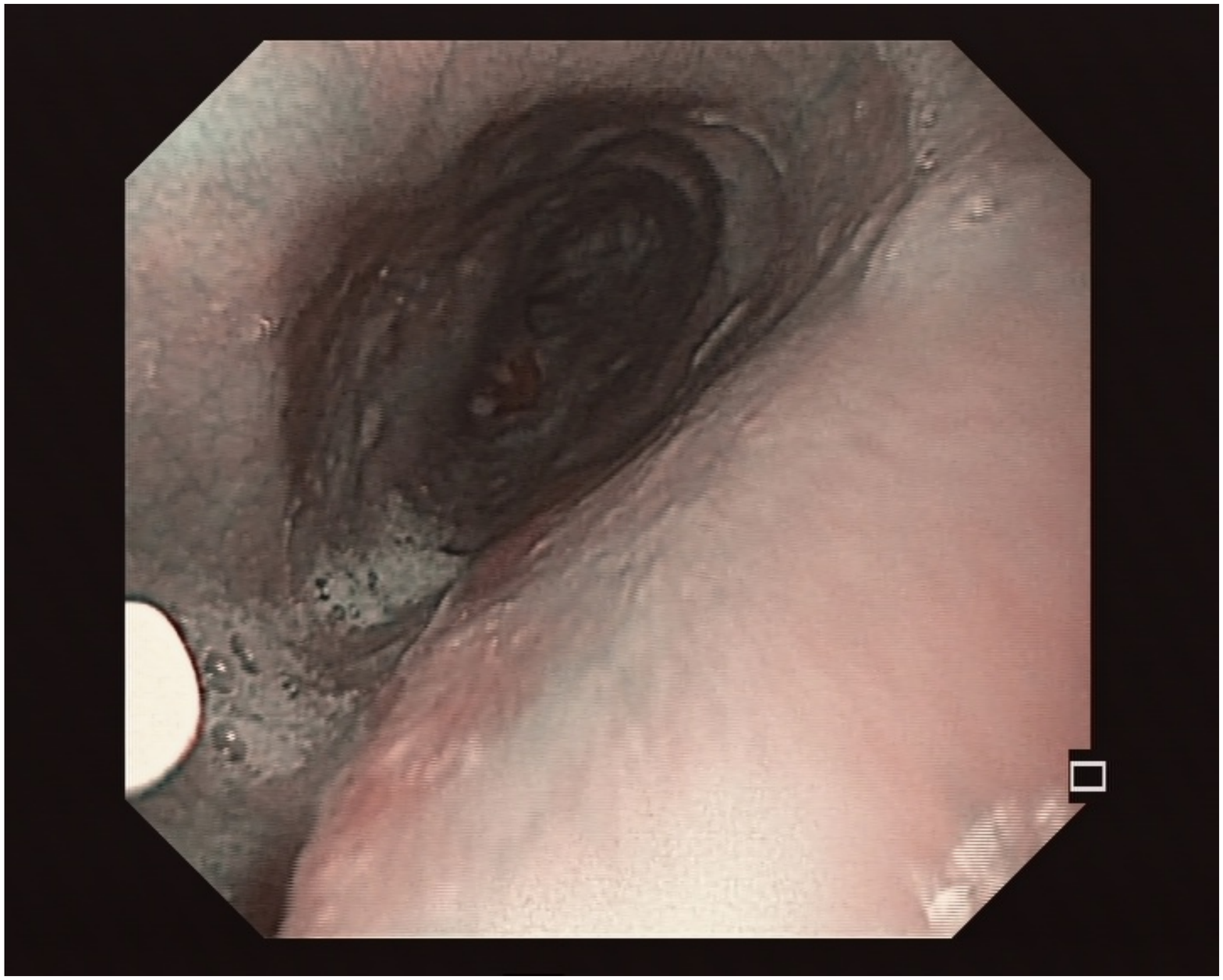}}
		\end{minipage}
		\begin{minipage}[h]{0.17\linewidth}
			 \centering
			 \centerline{\includegraphics[width=\linewidth]{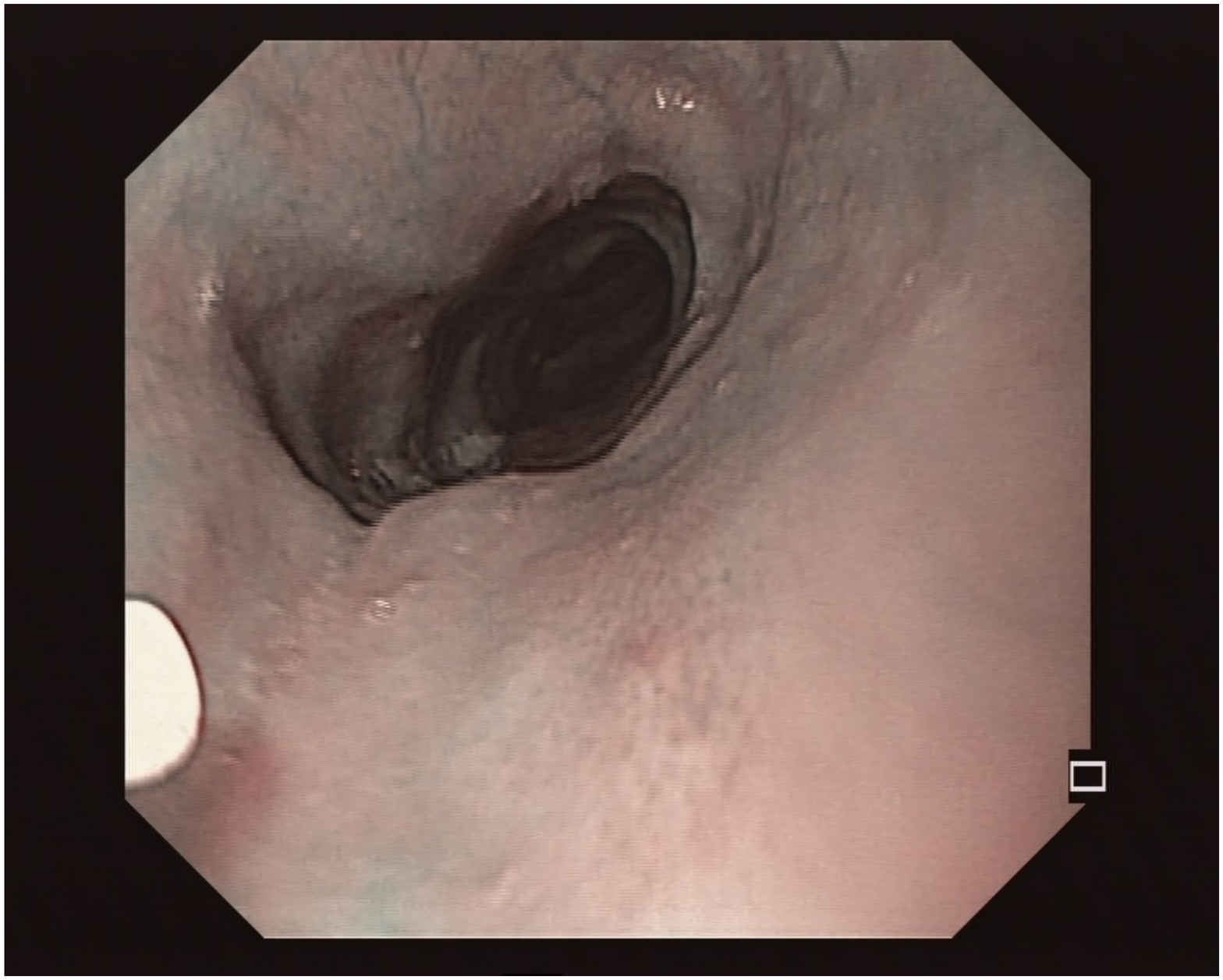}}
		\end{minipage}
		
		
		
		
		\begin{minipage}[h]{0.17\linewidth}
		 	\centering
		 	\centerline{\includegraphics[width=\linewidth]{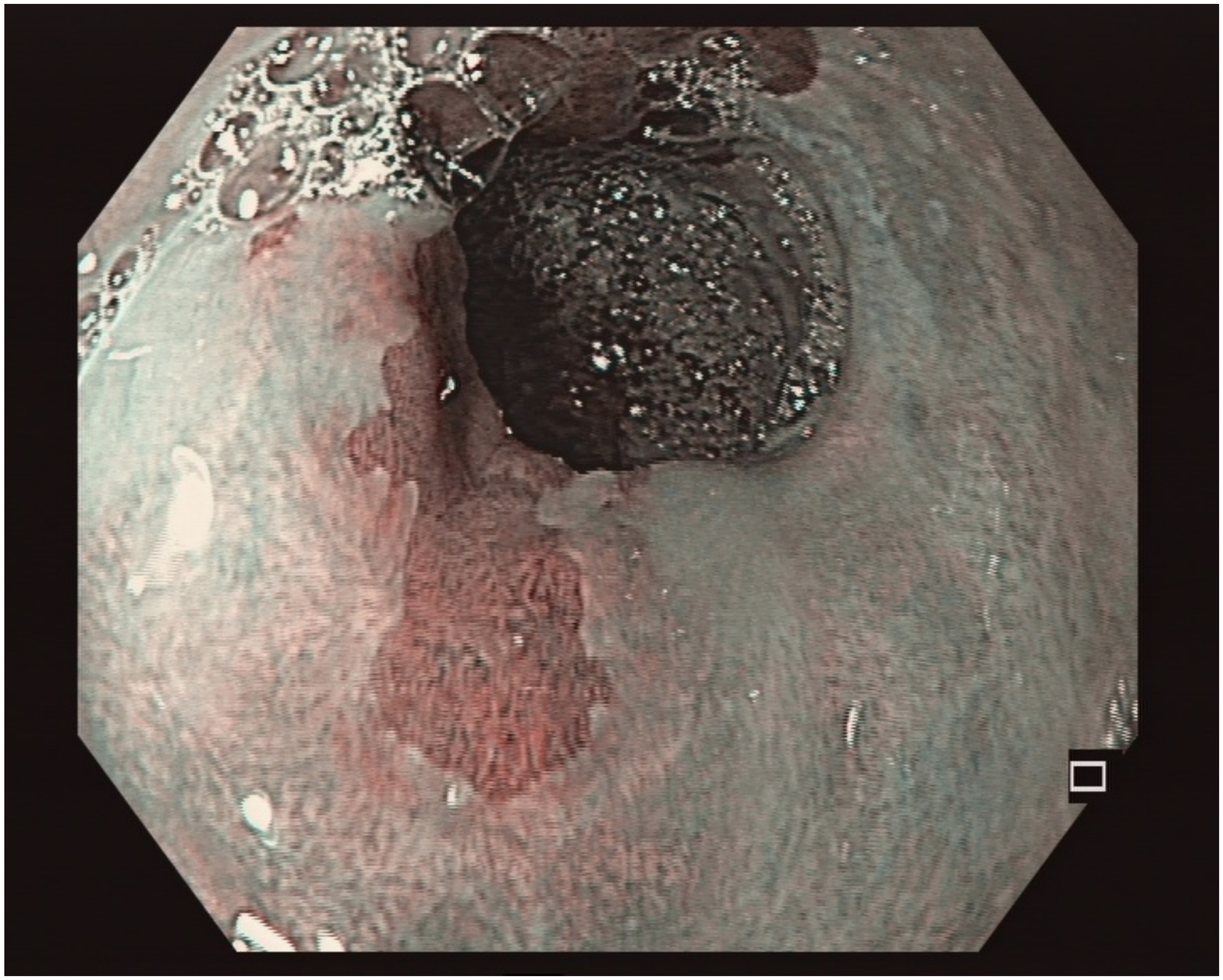}}
		\end{minipage}
		\begin{minipage}[h]{0.17\linewidth}
			 \centering
			 \centerline{\includegraphics[width=\linewidth]{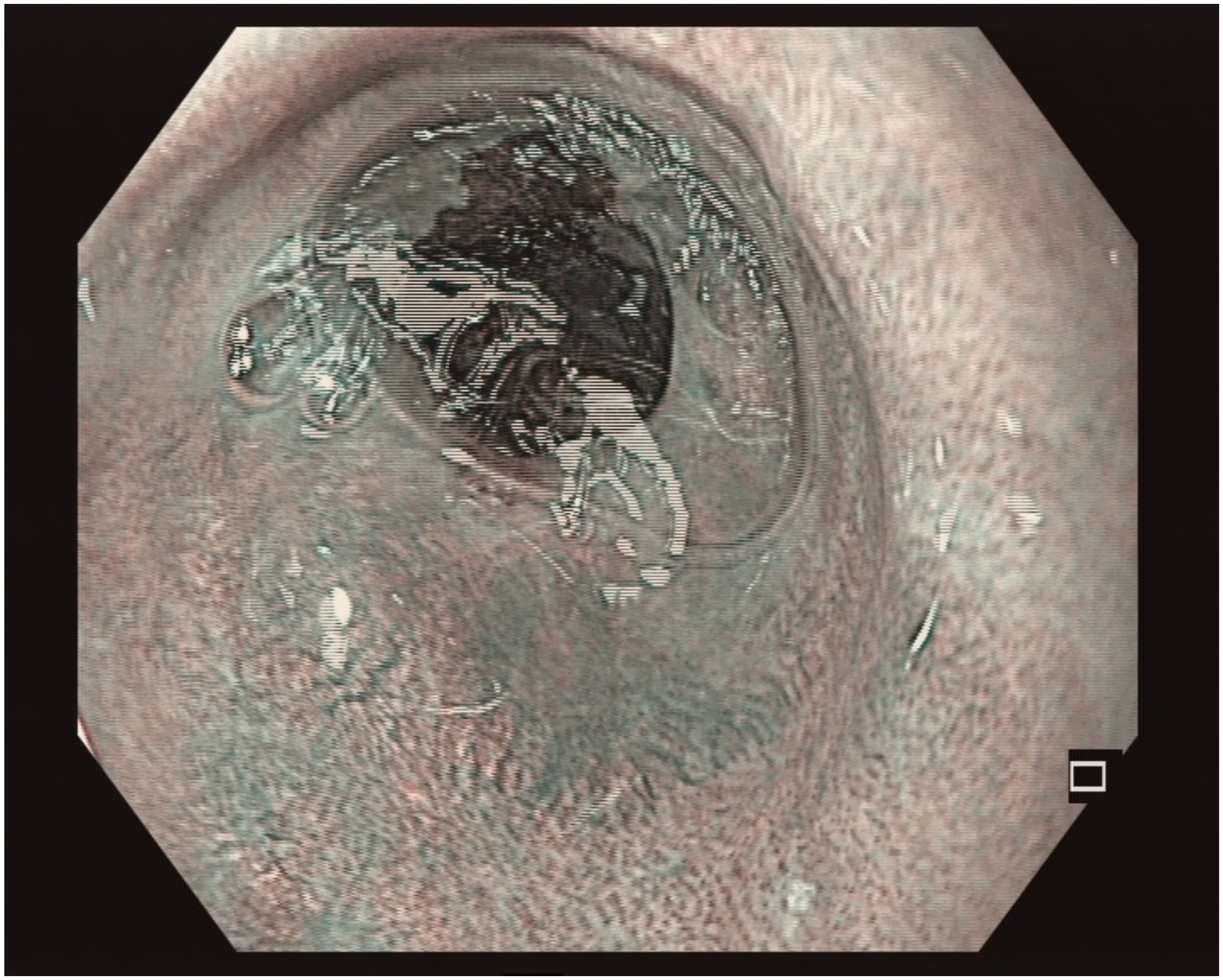}}
		\end{minipage}
		\begin{minipage}[h]{0.17\linewidth}
			 \centering
			 \centerline{\includegraphics[width=\linewidth]{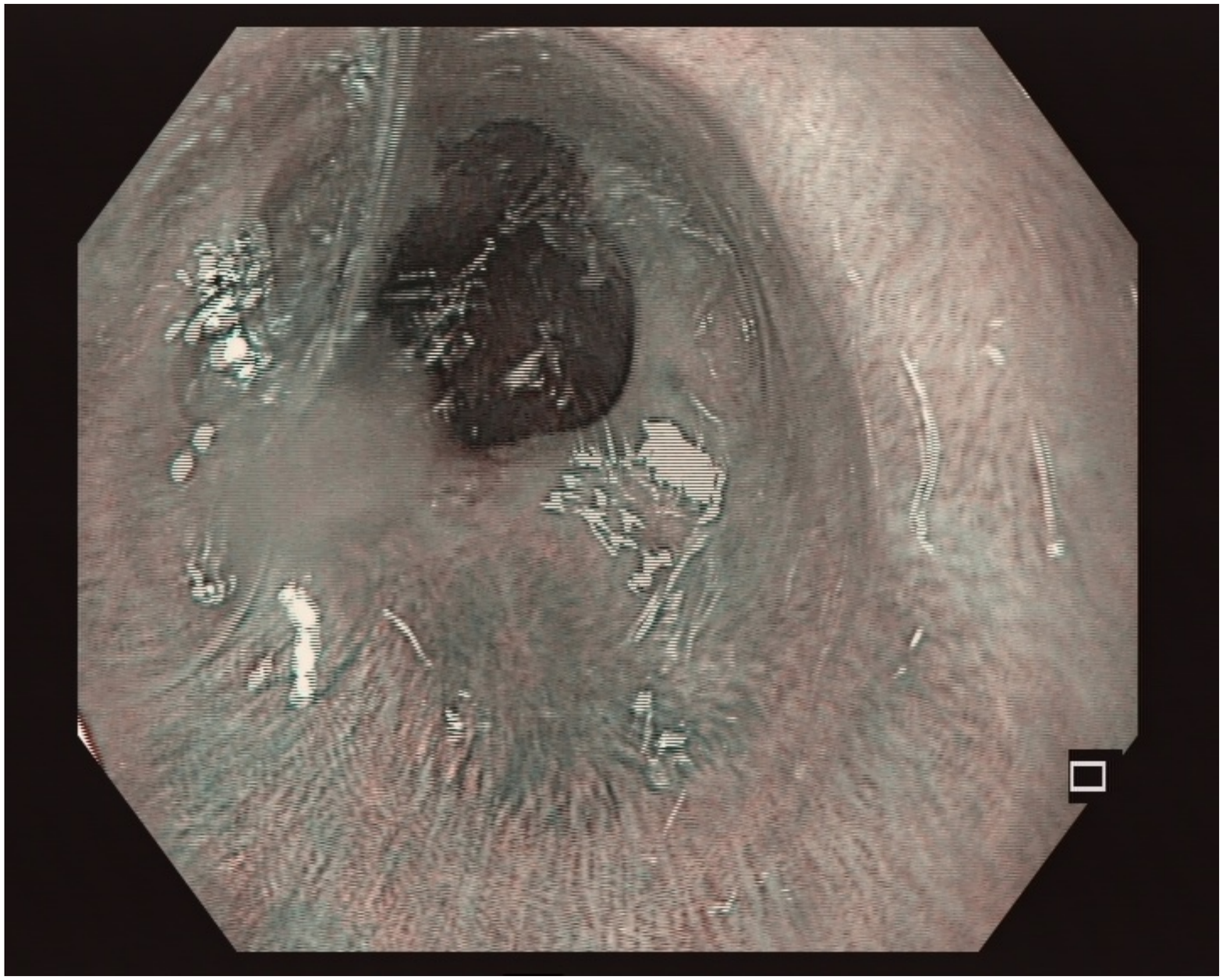}}
		\end{minipage}
		\begin{minipage}[h]{0.17\linewidth}
			 \centering
			 \centerline{\includegraphics[width=\linewidth]{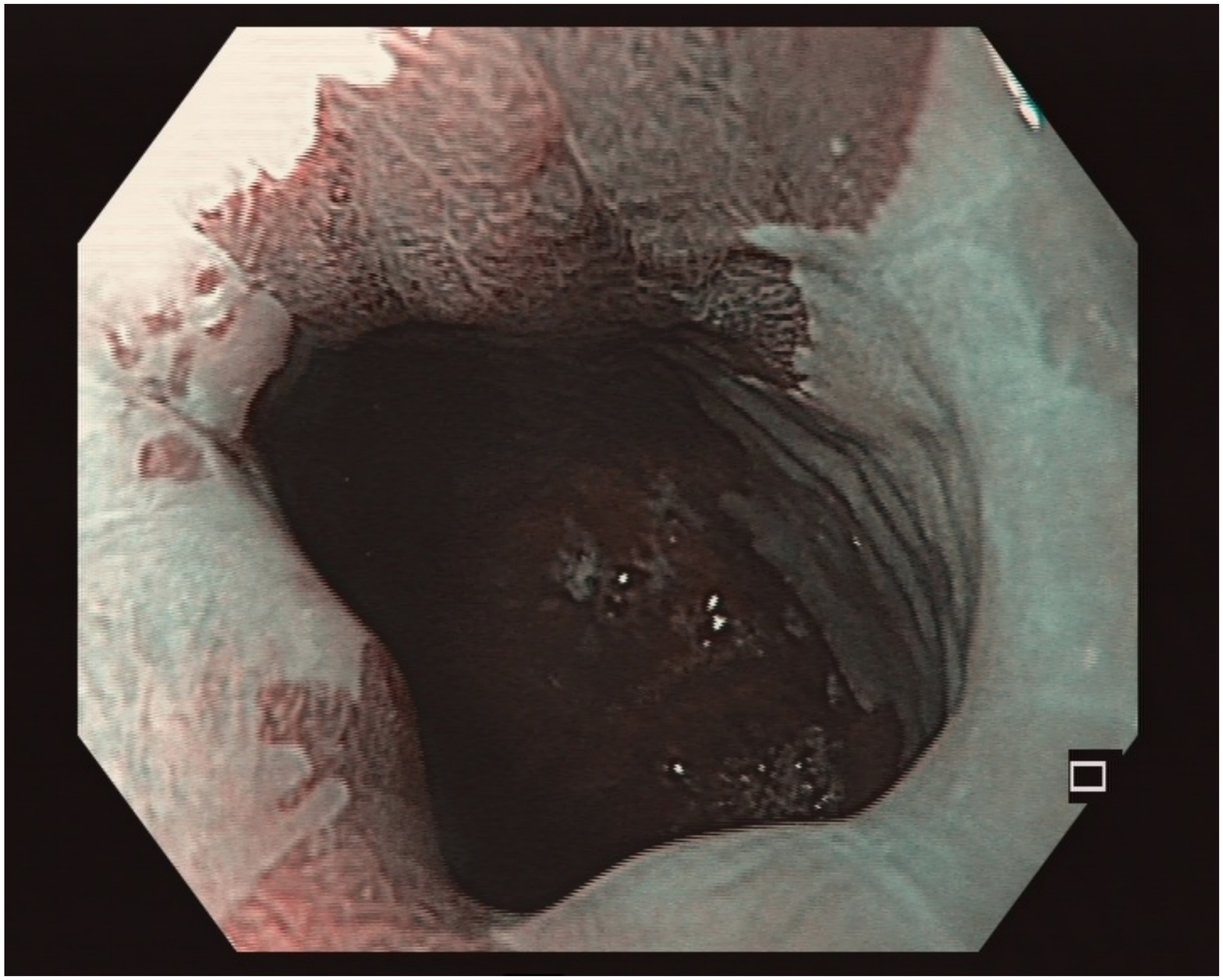}}
		\end{minipage}
		\begin{minipage}[h]{0.17\linewidth}
			 \centering
			 \centerline{\includegraphics[width=\linewidth]{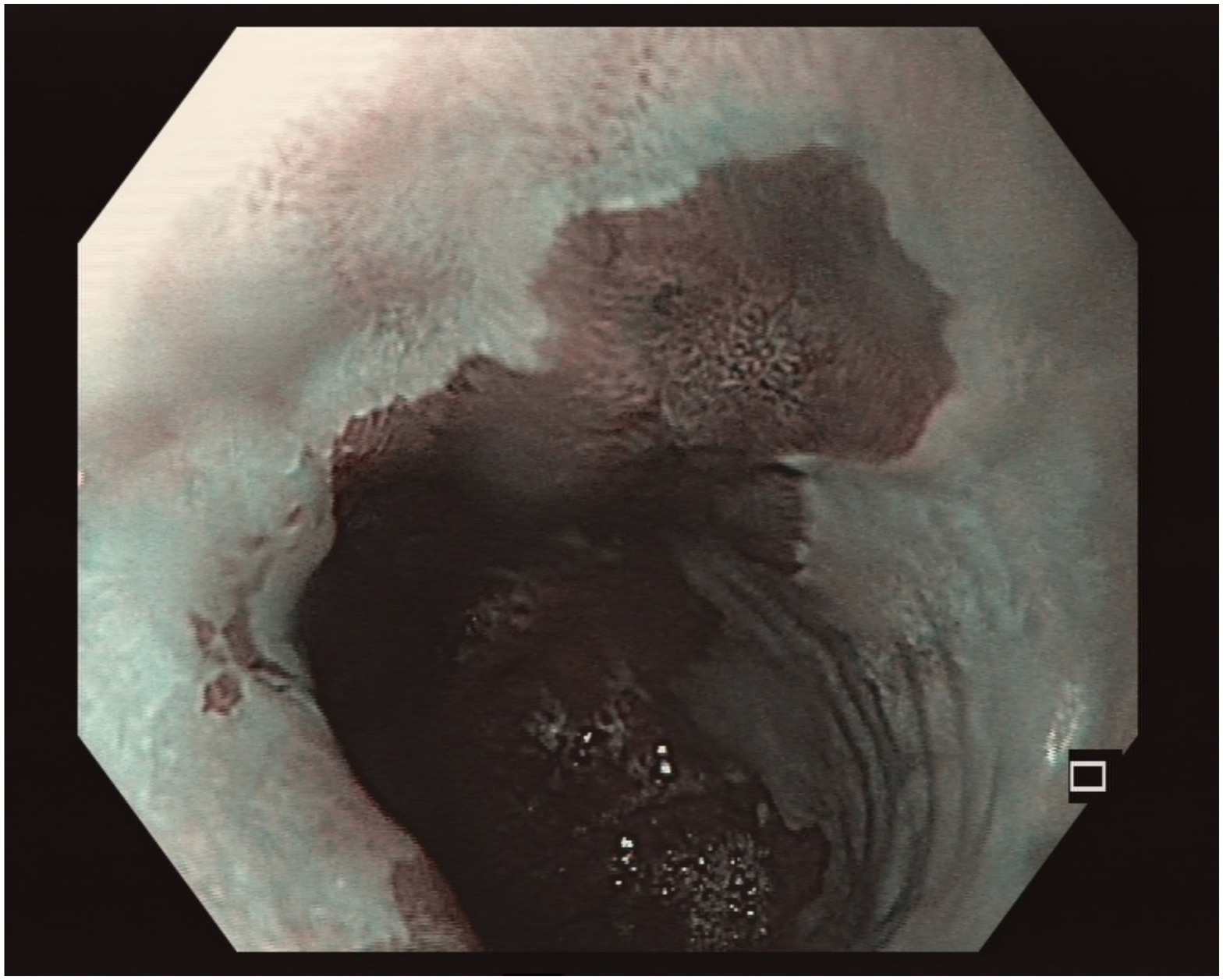}}
		\end{minipage}
		
		
		\begin{minipage}[h]{0.17\linewidth}
		 	\centering
		 	\centerline{\includegraphics[width=\linewidth]{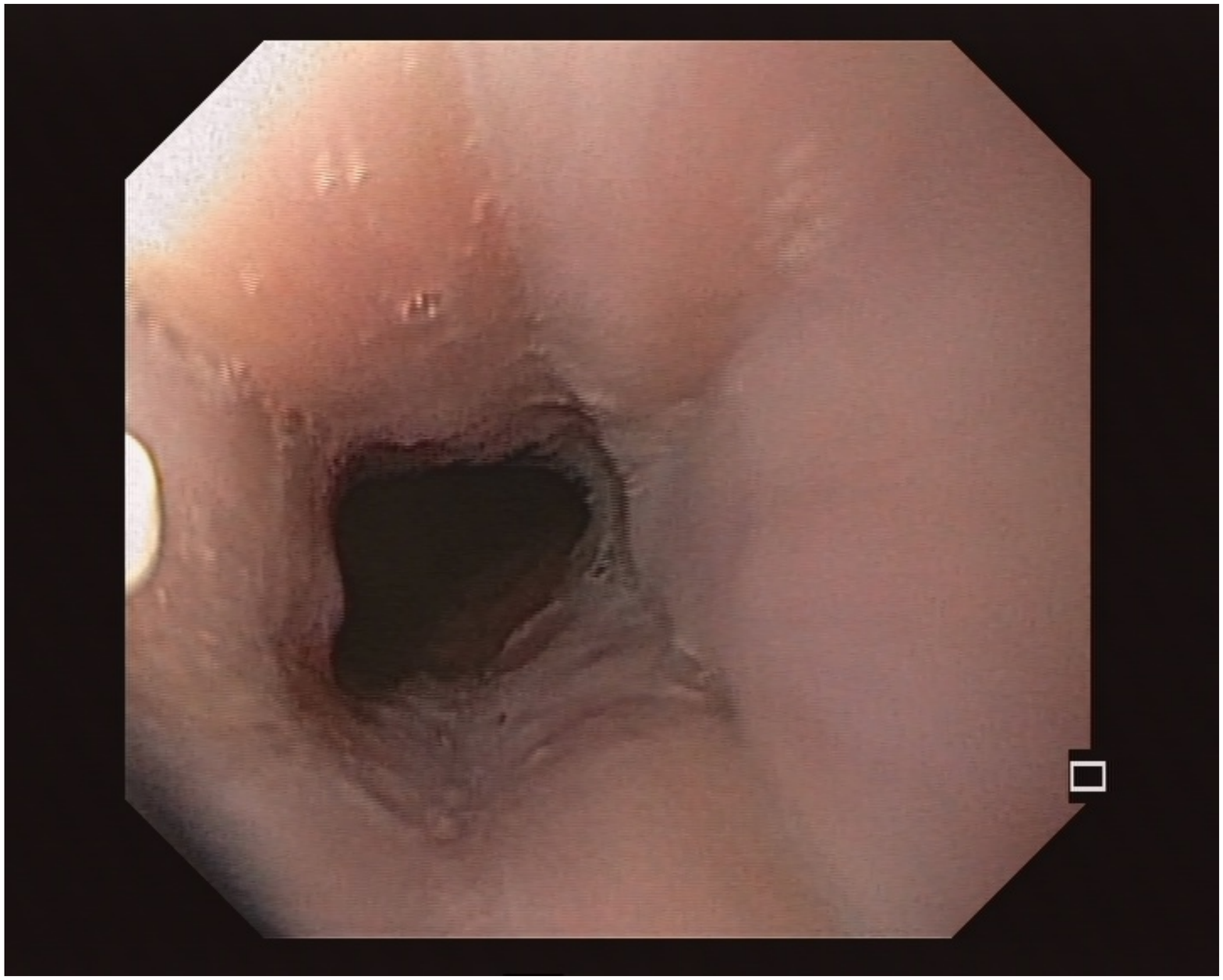}}
		\end{minipage}
		\begin{minipage}[h]{0.17\linewidth}
			 \centering
			 \centerline{\includegraphics[width=\linewidth]{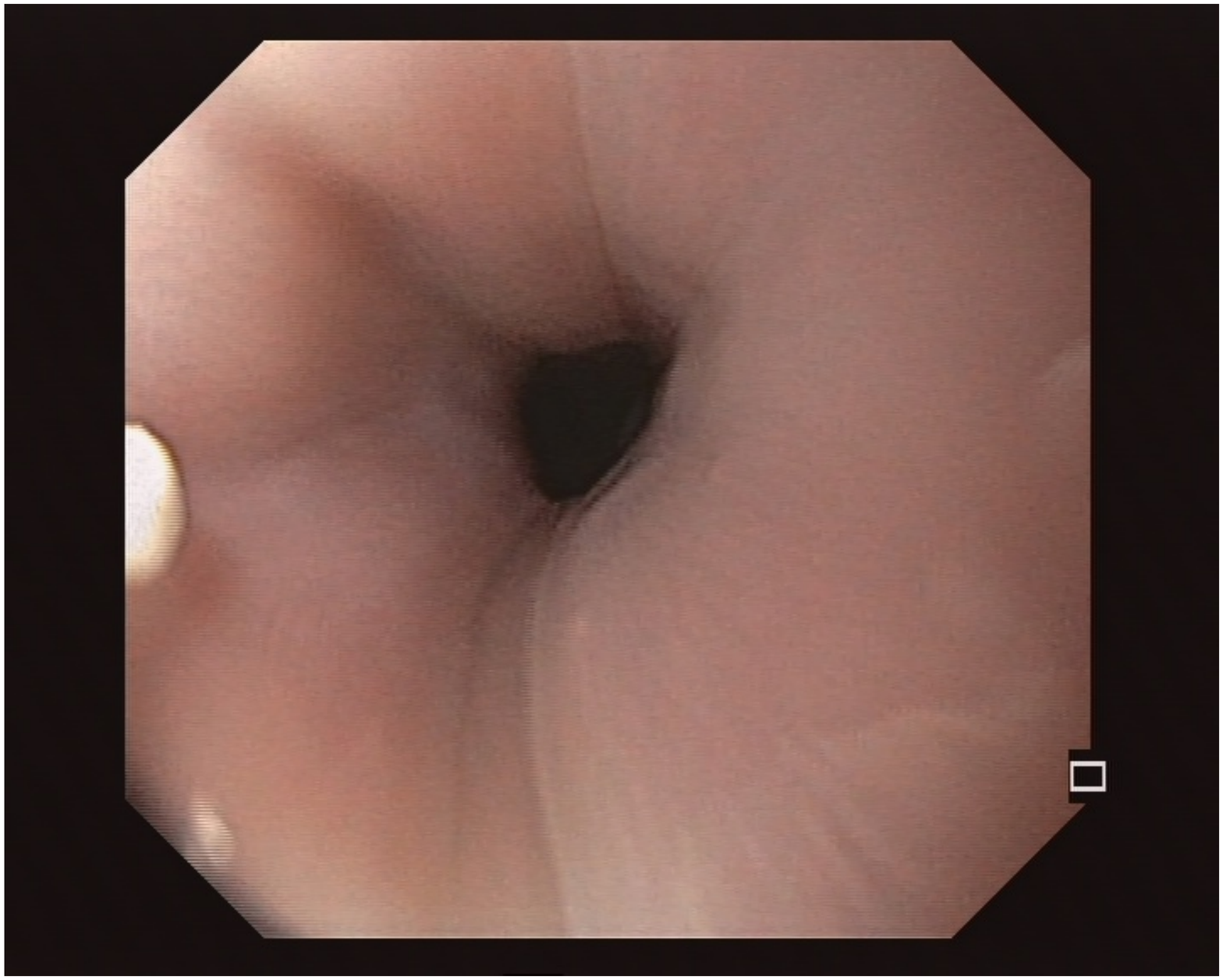}}
		\end{minipage}
		\begin{minipage}[h]{0.17\linewidth}
			 \centering
			 \centerline{\includegraphics[width=\linewidth]{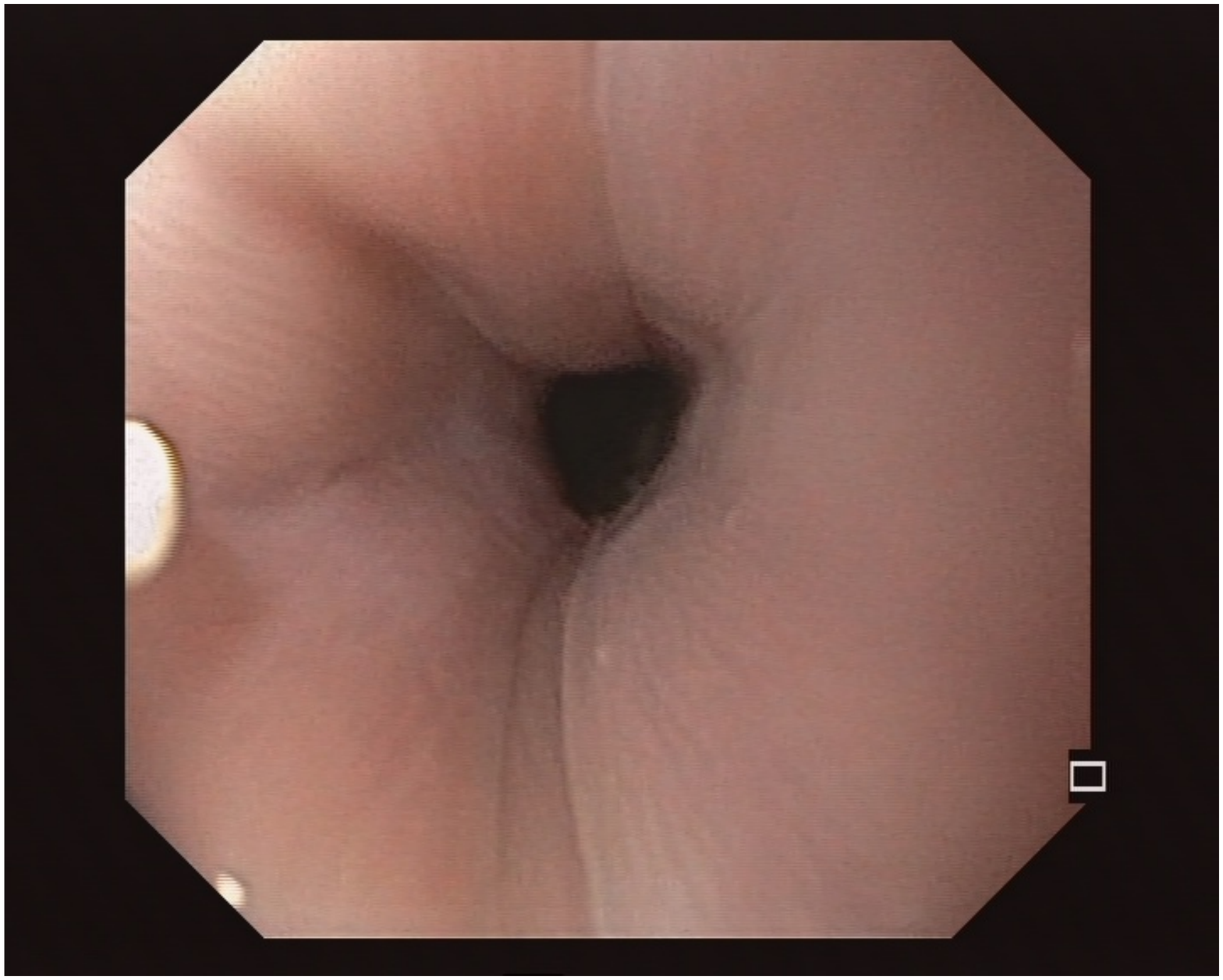}}
		\end{minipage}
		\begin{minipage}[h]{0.17\linewidth}
			 \centering
			 \centerline{\includegraphics[width=\linewidth]{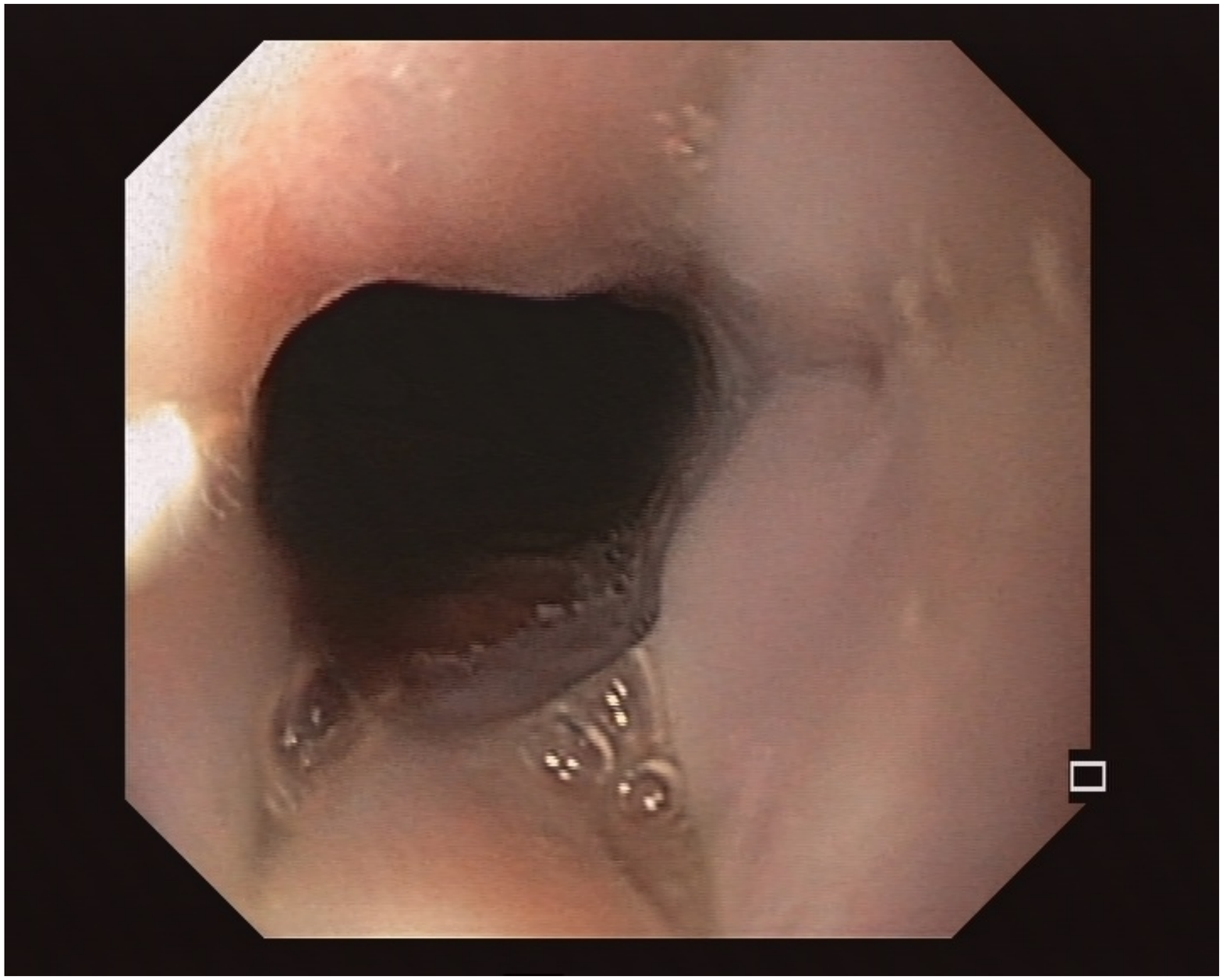}}
		\end{minipage}
		\begin{minipage}[h]{0.17\linewidth}
			 \centering
			 \centerline{\includegraphics[width=\linewidth]{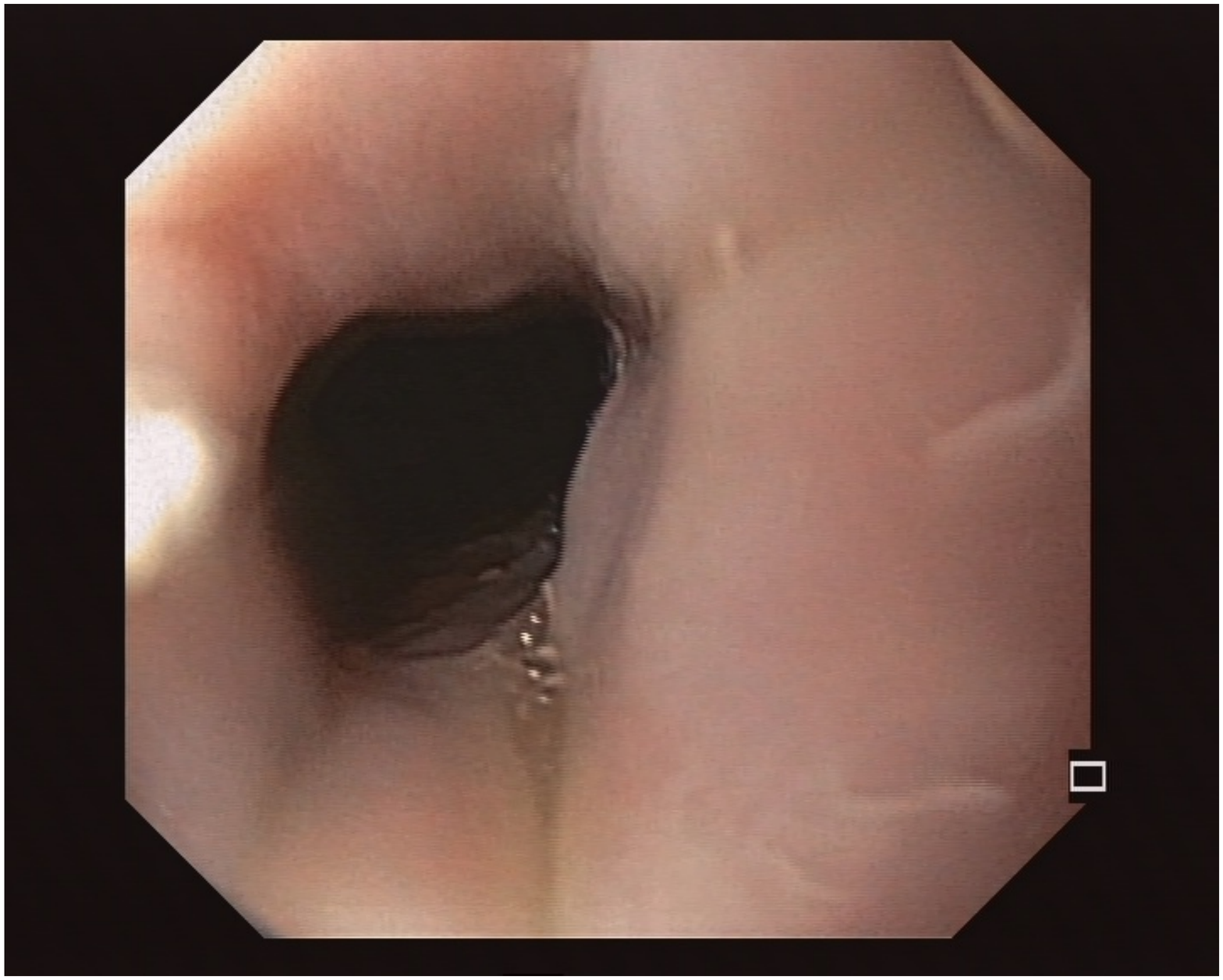}}
		\end{minipage}
		

		\begin{minipage}[h]{0.17\linewidth}
		 	\centering
		 	\centerline{\includegraphics[width=\linewidth]{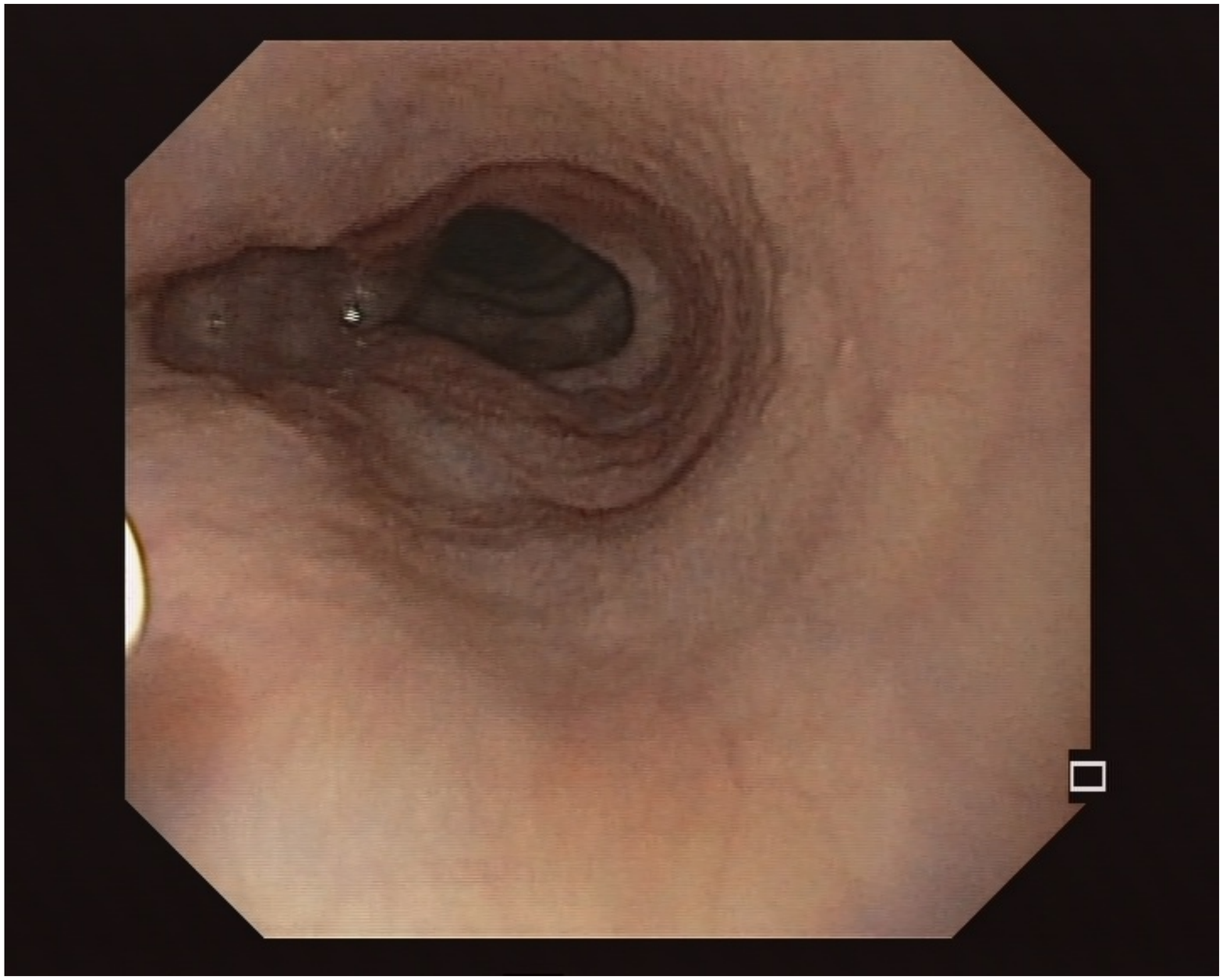}}
		\end{minipage}
		\begin{minipage}[h]{0.17\linewidth}
			 \centering
			 \centerline{\includegraphics[width=\linewidth]{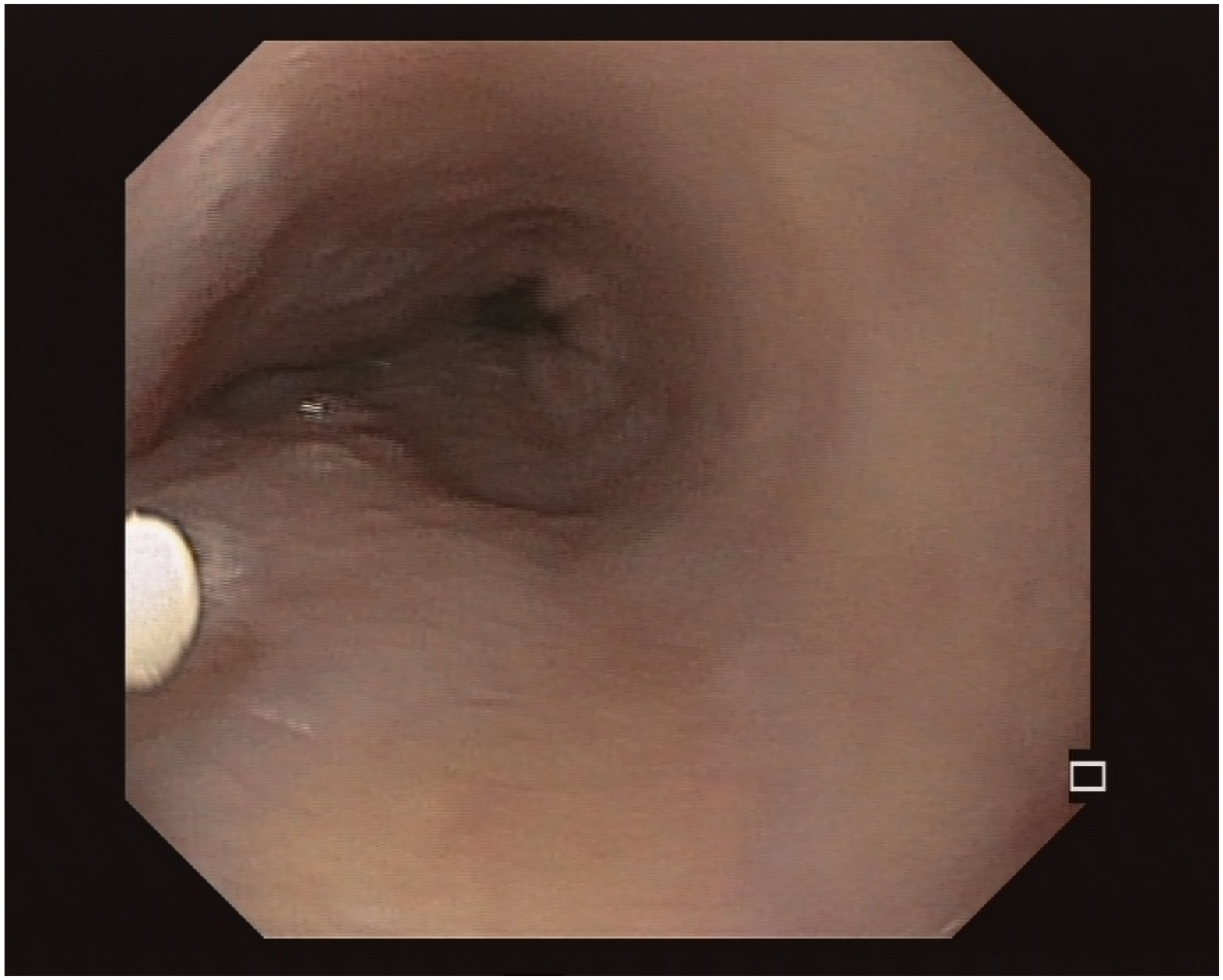}}
		\end{minipage}
		\begin{minipage}[h]{0.17\linewidth}
			 \centering
			 \centerline{\includegraphics[width=\linewidth]{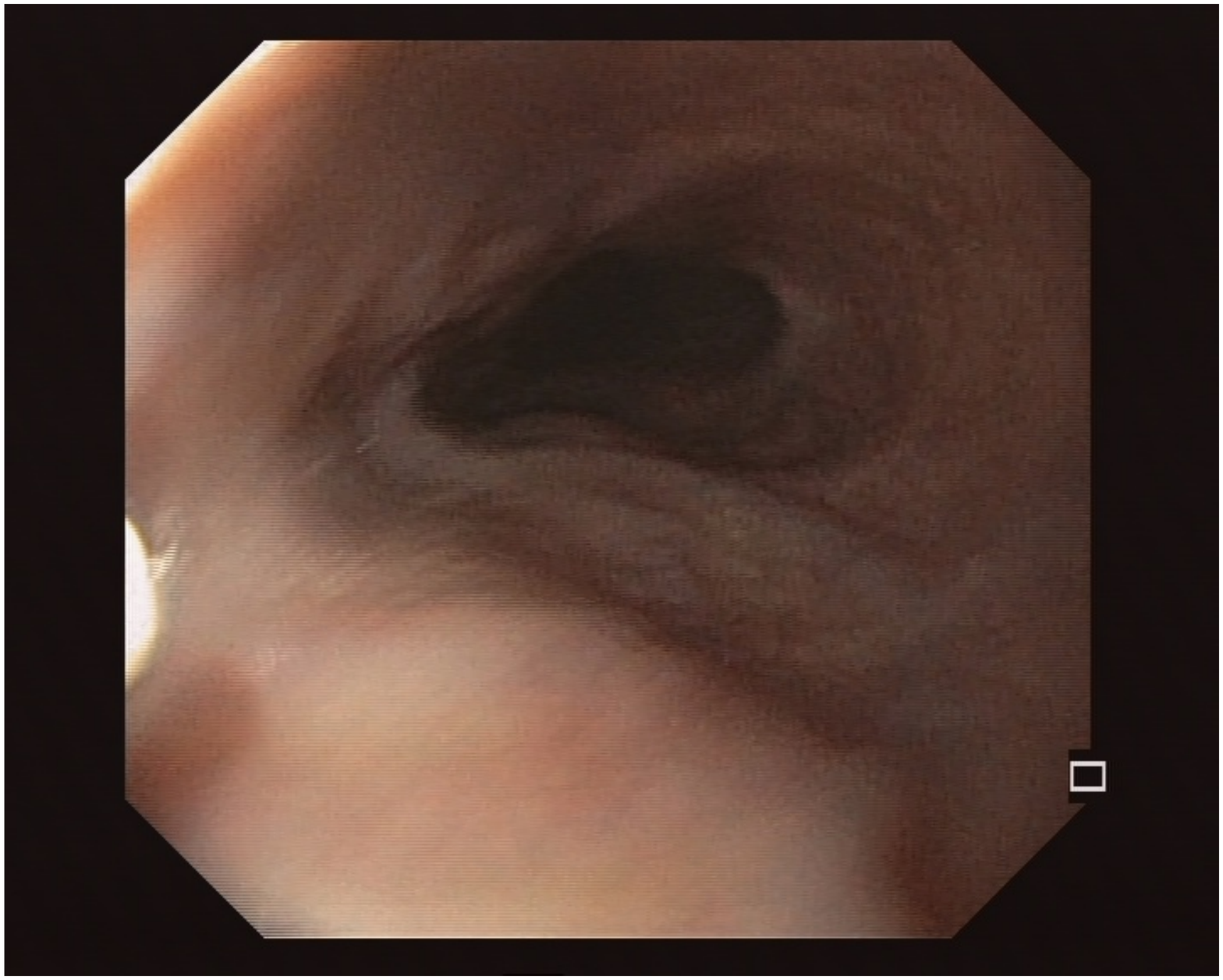}}
		\end{minipage}
		\begin{minipage}[h]{0.17\linewidth}
			 \centering
			 \centerline{\includegraphics[width=\linewidth]{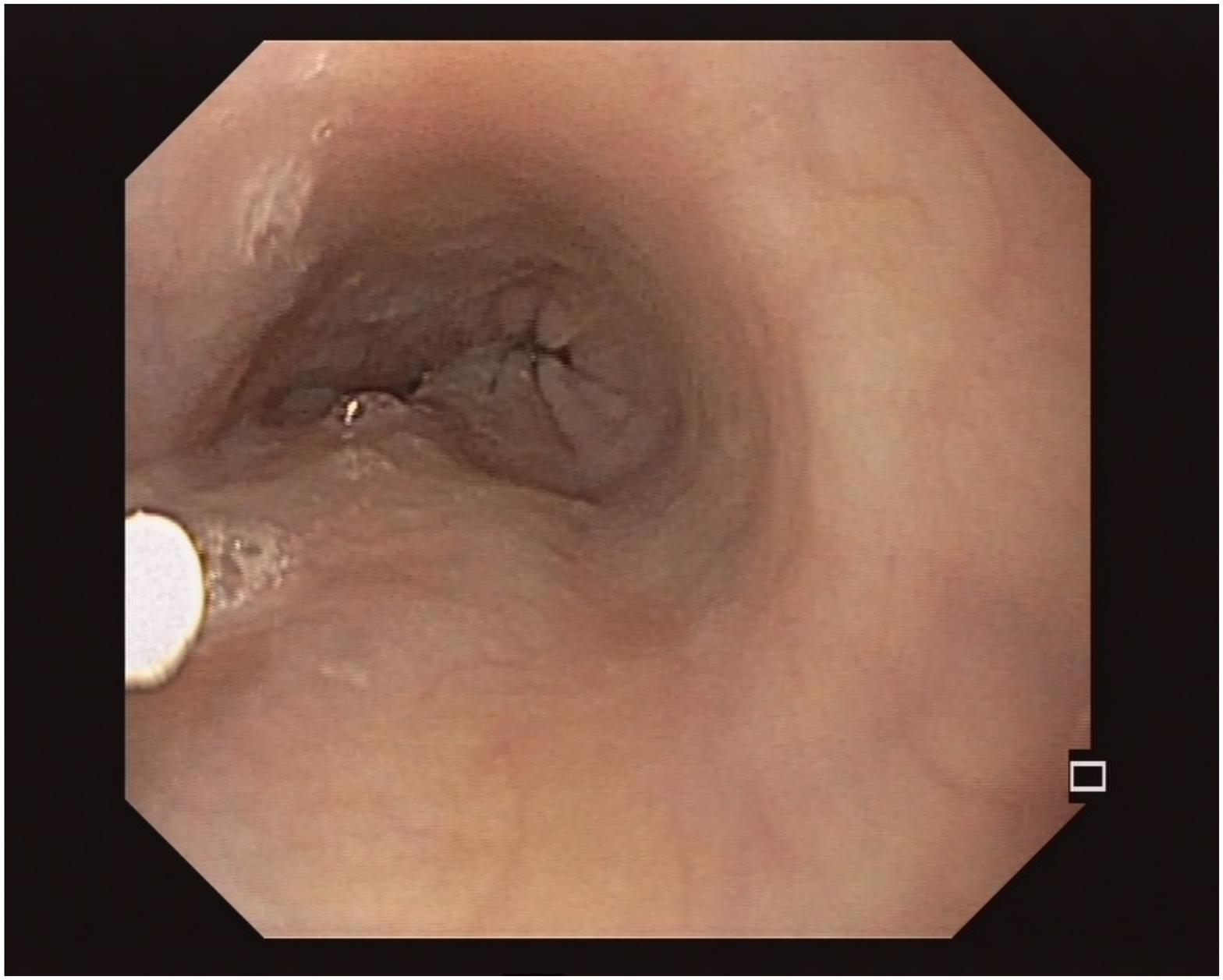}}
		\end{minipage}
		\begin{minipage}[h]{0.17\linewidth}
			 \centering
			 \centerline{\includegraphics[width=\linewidth]{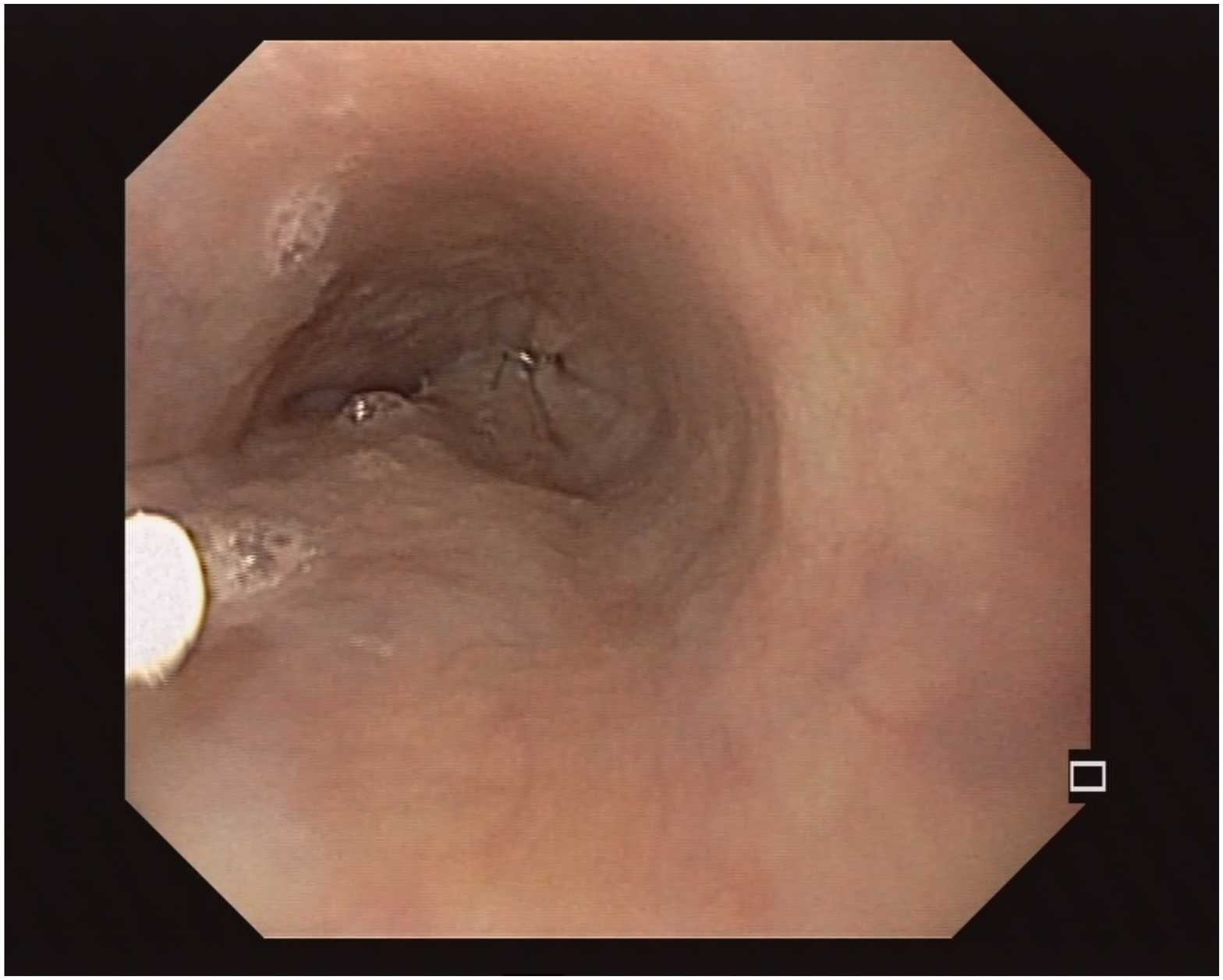}}
		\end{minipage}
		\caption{Column 1 : query frames. Columns 2 \& 3: the closest two EMNN matches. Columns 4 \& 5: the best view-point localized images. Scores for matches for each row ordered by their columns [2,3,4,5]. Row 1: [0,1,2,1], Row 2: [0,0,2,2], Row 3: [0,0,2,2], Row 4: [0,1,2,2]}
		\label{fig:results}
	\end{center}
\end{figure}

For training the classifier of UI frames, the GIS reviewed the images from 10 NBI and 8 WL surveillance interventions. A score of 2 for informative frame, 1 for partially informative and 0 for UI frame were assigned. For the purposes of this paper, we decided to combine the first two classes together. A total of 4236 NBI frames and 2643 WL frames were tagged. The precision and recall, for NBI = $[98\%,93\%]$ and WL = $[97\%,88\%]$. The average scores for EM based match for NBI and WL improved from $[0.97,0.82]$ to $[1.2,1.2]$ after filtering the UI frames. Finally, fig. (\ref{fig:results}) shows the best matches from EM and imaged based approach. 


\section{Discussion and Conclusion}
\label{sec:discussion_conclusion}
\begin{figure}[!htp]
      \centering
      \begin{minipage}[h]{0.49\linewidth}
	      \centering
	      \includegraphics[trim=2.2cm 7.2cm 1.8cm 9.2cm,width=0.92\linewidth]{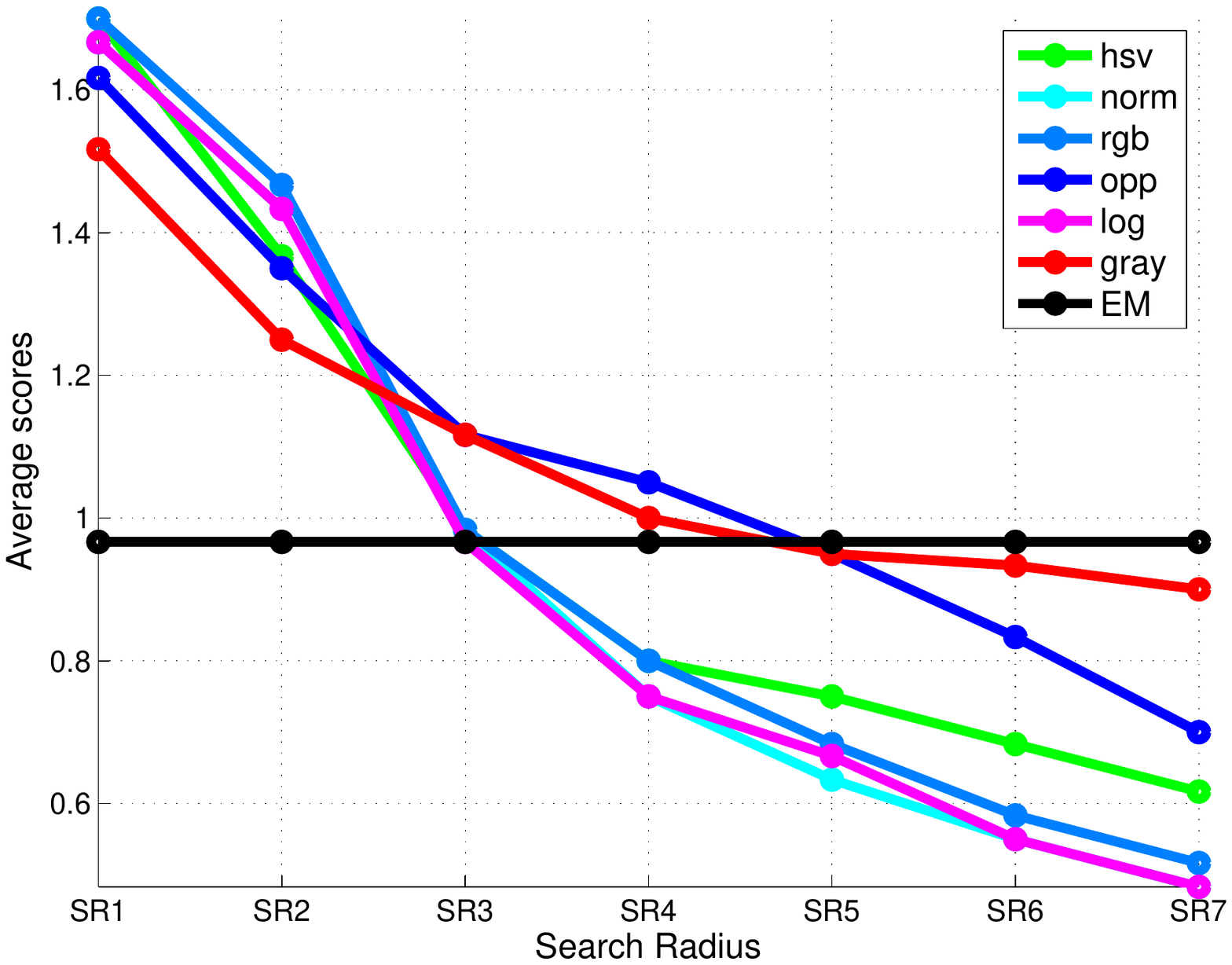}
      \end{minipage}
      \begin{minipage}[h]{0.49\linewidth}
	      \centering
	      \includegraphics[trim=2.2cm 7.2cm 1.8cm 9.2cm,width=0.92\linewidth]{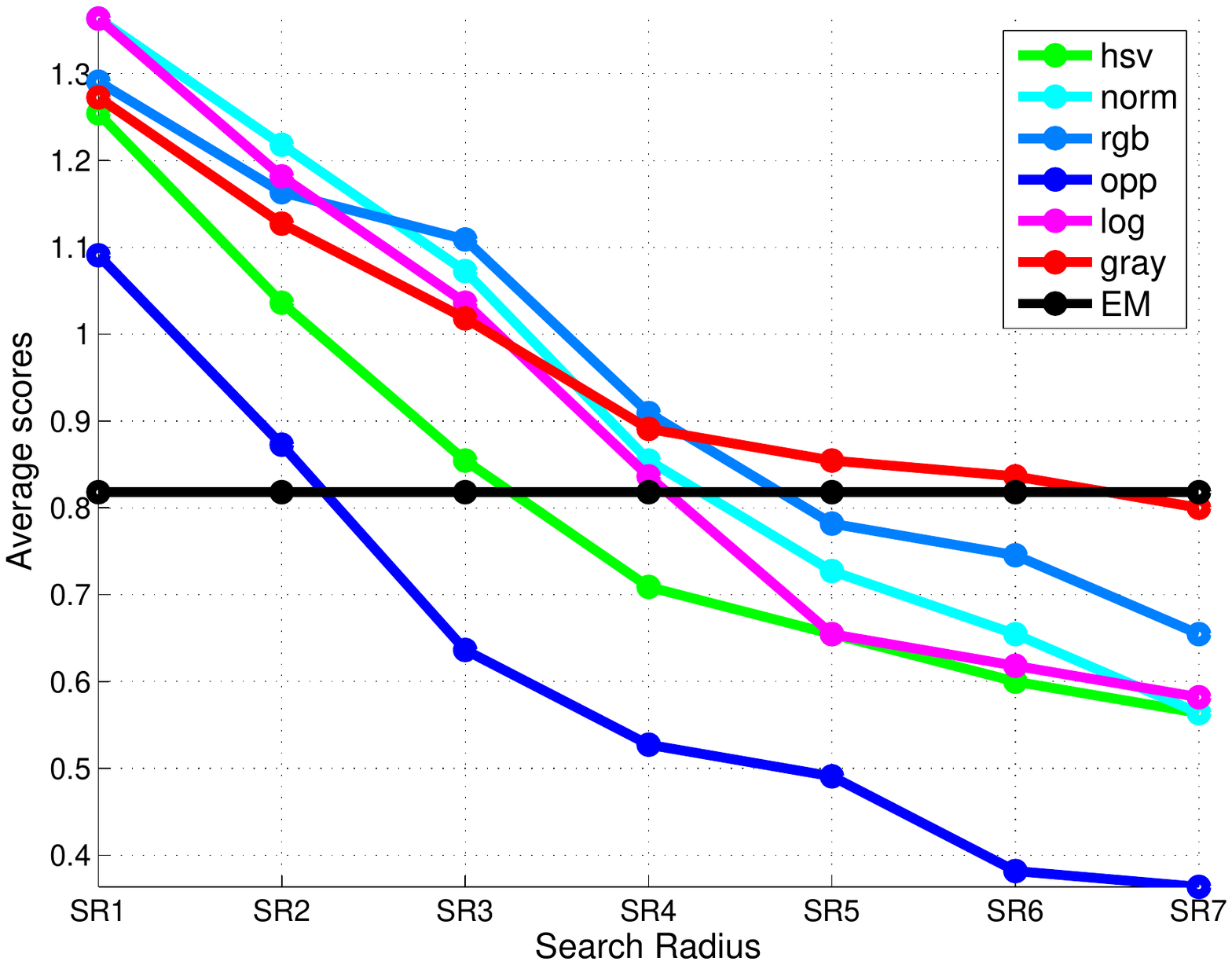}
      \end{minipage}
      \caption{Avg. scores of mLBP descriptor for six color-spaces over increasing search radii (10mm to 70mm). NBI (left) and WL (right).}
      \label{fig:score_plot_sr1_sr7}
\end{figure}

\begin{figure}[!htb]
      \centering
      \begin{minipage}[h]{0.49\linewidth}
	      \centering
	      \includegraphics[trim=2.2cm 7.2cm 1.8cm 8cm,width=0.92\linewidth]{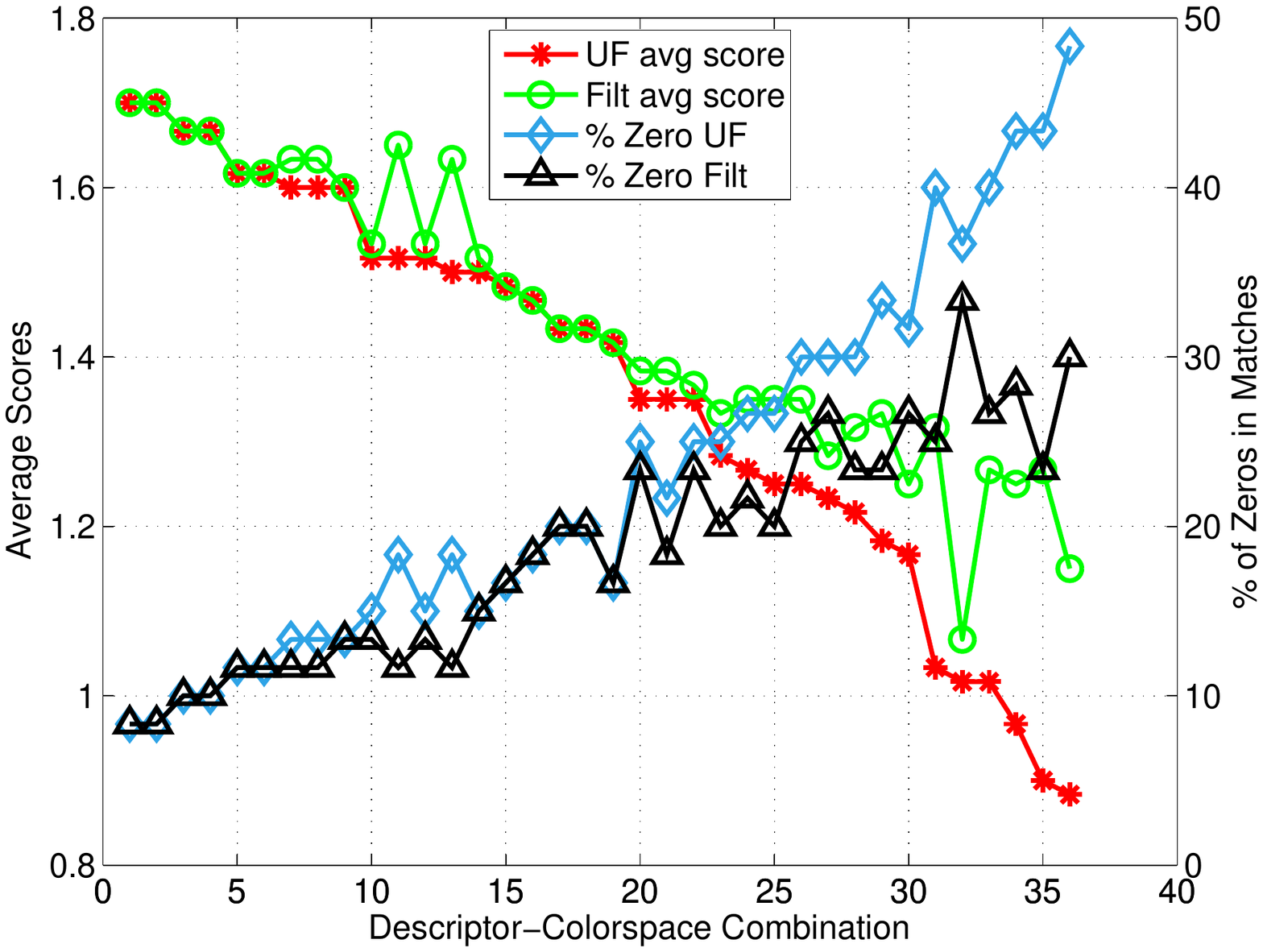}
      \end{minipage}
      \begin{minipage}[h]{0.49\linewidth}
	      \centering
	      \includegraphics[trim=2.2cm 7.2cm 1.8cm 8cm,width=0.92\linewidth]{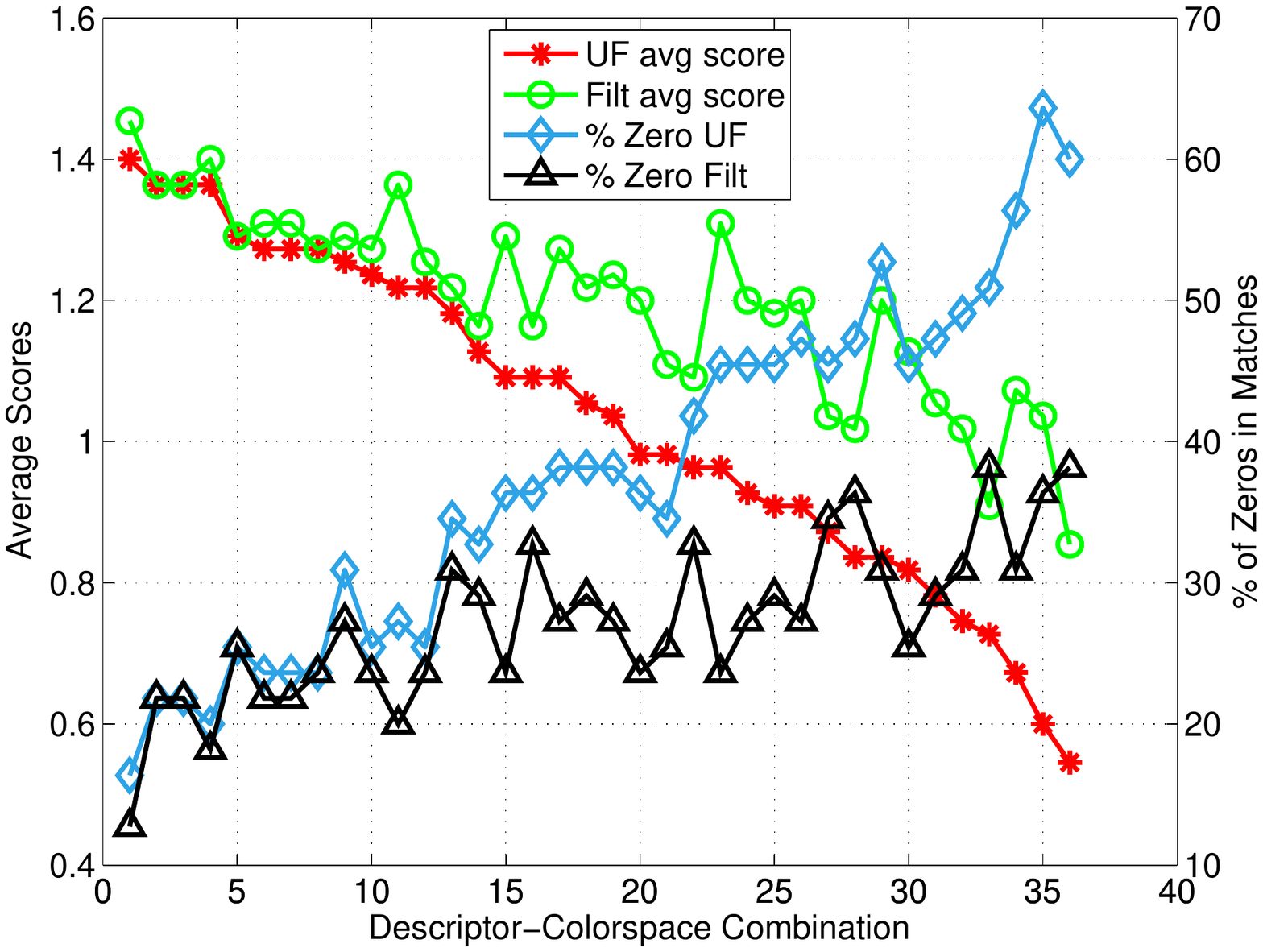}
      \end{minipage}
      \caption{Avg. scores on the left axis and \% of images matched with score zero on the right axis for all 36 of the descriptor-color-space combinations. The x-axis is sorted by the best avg. score. NBI (left) and WL (right).}
      \label{fig:score_plot_sr1}
\end{figure}

This paper extends our earlier work on inter-operative synchronization to include an important aspect of view-point localization. We compared various descriptors used for texture and scene classification discussed in the literature and also presented an alternative approach to UI frame removal. To our knowledge this is the first paper that provides such a comparison over different endoscopic modalities. Most importantly, the presented framework allows for quantitative evaluation of inter-operative view-point matching which is an important aspect of differential surveillance. We observed that, \begin{inparaenum}[\upshape(i\upshape)]
\item the performance is better on NBI than on WL, which is expected because of the higher texture in NBI modality.
\item Fig. (\ref{fig:score_plot_sr1_sr7}) depicts the general trend of decreasing avg. score with increasing search radius, indicating the need for constraining the search space using gross-localization, whereas in \cite{Atasoy2012} only temporal localization of frames was considered.
\item Fig. (\ref{fig:score_plot_sr1}) clearly shows that filtering the uninformative frames reduces the number of false matches. This is however more observable in WL than in NBI.
\item Using GS for NBI images is not meaningful as observed from Table \ref{tab:NBI_WL_FilteredScores_SR1}, because it is not a true mapping from RGB space.
\item Table \ref{tab:NBI_WL_FilteredScores_SR1} also shows that texture based descriptors such as LBP, LIOP and LTP are much better suited in this scenario. Along with the choice of illumination invariant color-spaces such as hsv, norm and log; variations of these must be explored further.
\end{inparaenum}

\begin{table}[ht]
    \centering
    \resizebox{0.875\textwidth}{!}{
	\begin{tabular}{llllllll}
	    \toprule
	    \centering
	    \emph{\small{Modality}} & \emph{\small{Descriptor}} & \emph{\small{color-space}} & \emph{\small{Avg. Score}} & \emph{\small{Std. dev.}} & \emph{\small{\% Zeros}} & \emph{\small{\% Ones}} & \emph{\small{\% Twos}}\\
	    \midrule
	    \multirow{10}{*}{NBI}& mLBP & hsv & 1.7 & 0.619 & 8.33 & 13.33 & 78.33\\
	    & mLBP & rgb & 1.7 & 0.619 & 8.33 & 13.33 & 78.33\\
	    & mLBP & norm & 1.667 & 0.655 & 10 & 13.33 & 76.67\\
	    & mLBP & log & 1.667 & 0.655 & 10 & 13.33 & 76.67\\
	    & mLBP & OPP & 1.617 & 0.691 & 11.67 & 15 & 73.33\\
	    & mLBPHOG & hsv & 1.65 & 0.685 & 11.67 & 11.67 & 76.67\\
	    & mLBPHOG & norm & 1.633 & 0.688 & 11.67 & 13.33 & 75\\
	    & mLBPHOG & rgb & 1.633  & 0.688 & 11.67 & 13.33 & 75\\
	    & mLBPHOG & log & 1.633  & 0.688 & 11.67 & 13.33 & 75\\
	    \rowcolor{yellow}& EM-Based & n.a. & 1.15 & 0.82 & 26.67 & 31.67 & 41.67\\
	    \midrule
	    \multirow{10}{*}{WL}& LIOP & gs & 1.455 & 0.715 & 12.73 & 29.09 & 58.18\\
	    & swmLBP & gs & 1.4 & 0.784 & 18.18 & 23.64 & 58.18\\
	    & mLTP & gs & 1.364 & 0.802 & 20 & 23.64 & 56.36\\
	    & mLBP & norm & 1.364 & 0.825 & 21.82 & 20 & 58.18\\
	    & mLBP & log & 1.364 & 0.825 & 21.82 & 20 & 58.18\\
	    & mLBP & gs & 1.309 & 0.814 & 21.82 & 25.45 & 52.73\\
	    & mLBPHOG & gs & 1.309 & 0.814 & 21.82 & 25.45 & 52.73\\
	    & mLBP & OPP & 1.291 & 0.832 & 23.64 & 23.64 & 52.73\\
	    & mHOG & gs & 1.2 & 0.803 & 23.64 & 32.73 & 43.64\\
	    \rowcolor{yellow}& EM-Based & n.a. & 1.182 & 0.796 & 23.64 & 34.55 & 41.82\\
	    \bottomrule
	\end{tabular}}
	\caption{Avg. scores of descriptors-color-space combinations for NBI and WL (filtered of UI frames), sorted in the increasing order of column 6 (truncated to 9 best values). Columns 6,7 and 8 are the \% of cases where the matched image has a score of zero, one and two respectively.}
	\label{tab:NBI_WL_FilteredScores_SR1}
\end{table}

\bibliographystyle{unsrt}
\bibliography{\jobname}

\end{document}